\documentclass[journal]{IEEEtran}
\usepackage{amsmath,epsfig}
\usepackage{paralist,amssymb, amsfonts}
\usepackage{algcompatible}
\usepackage{algorithm}
\usepackage{cite}
\usepackage{tabularx}
\usepackage{setspace}
\usepackage{multirow}
\usepackage{bm}

\DeclareMathOperator*{\argmin}{arg\,min}
\DeclareMathOperator*{\argmax}{arg\,max}
\algnewcommand\algorithmicreturn{\textbf{return }}
\algnewcommand\RETURN{\State \algorithmicreturn}%

\begin{document}
\title{Exploiting Restricted Boltzmann Machines and Deep Belief Networks in Compressed Sensing}
\author{Luisa~F.~Polan\'{i}a,~\IEEEmembership{Member,~IEEE,}
        and~Kenneth~E.~Barner,~\IEEEmembership{Fellow,~IEEE}% <-this % stops a space
\thanks{L.F. Polan\'{i}a is with American Family Mutual Insurance Company, Madison, WI, 53783 USA (e-mail: lpolania@amfam.com).}
\thanks{K.E. Barner is with the Department
of Electrical and Computer Engineering, University of Delaware, Newark, DE, 19711 USA (e-mail: barner@udel.edu).}

}

\markboth{}%
{Shell \MakeLowercase{\textit{et al.}}: Bare Demo of IEEEtran.cls for Journals}

\maketitle

\begin{abstract}
 This paper proposes a CS scheme that exploits the representational power of restricted Boltzmann machines and deep learning architectures to model the prior distribution of the sparsity pattern of signals belonging to the same class. The determined probability distribution is then used in a maximum \textit{a posteriori} (MAP) approach for the reconstruction. The parameters of the prior distribution are learned from training data. The motivation behind this approach is to model the higher--order statistical dependencies between the coefficients of the sparse representation, with the final goal of improving the reconstruction. The performance of the proposed method is validated on the Berkeley Segmentation Dataset and the MNIST Database of handwritten digits.
\end{abstract}


\section{Introduction}
\IEEEPARstart{C}OMPRESSED sensing has become an extensive research area due to its potential to perfectly reconstruct sparse signals from a small set of nonadaptive linear measurements in the form of random projections~\cite{Dono06,Cand08}. In essence, CS states that data acquisition with far fewer measurements than that dictated by the Shannon-Nyquist theorem is possible, under certain conditions. In last decade, the area of CS has extended to new applications that require structured signal models that go beyond the simplistic sparsity model~\cite{Duar11, Elda10, Ji08, Seeg08, Carr10}. Examples of deterministic models include the wavelet tree model, which assumes that the non-zero signal coefficients lie in a rooted and connected tree structure, and the block-sparsity model, which assumes that the non-zero signal coefficients form clusters~\cite{Duar11, Elda10}. Instead of imposing an explicit structure of the coefficients, statistical approaches usually impose a prior belief about the signal of interest in terms of a sparseness prior~\cite{Ji08, Seeg08}.

Even though the bulk of CS theory has been developed for signals that have a sparse representation in an orthonormal basis, efforts have been made to extend CS theory to signals that are sparse with respect to an overcomplete dictionary~\cite{Rauh08, Elad07, Duar09}. This extension adds more flexibility to CS as many signals of interest are not sparse in an orthonormal basis, but are in an overcomplete dictionary. For example, reflected radar and sonar signals have a sparse representation in Gabor frames. The coherence between the columns of an overcomplete dictionary poses some limitations in extending the CS theory to sparse overcomplete representations~\cite{Dono06, Rauh08}. However, Raught \textit{et al}.~\cite{Rauh08} showed that CS is viable in the context of signals that are sparse in an overcomplete dictionary. They studied the conditions on the overcomplete dictionary that, in combination with a random sampling matrix, results in small restricted isometry constants.

In this paper, a statistical approach is proposed that uses restricted Boltzmann machines (RBMs) and deep belief networks (DBNs) to model the prior distribution of the sparsity pattern of the signal to be recovered. The proposed method requires \textit{a priori} training data of the same class as the signal of interest. Either orthonormal bases, such as the wavelet transform, or overcomplete learned dictionaries can be employed as sparsifying transforms in the proposed approach. In the case of overcomplete dictionaries, a training stage is employed with dual purpose. First, it learns an overcomplete dictionary to sparsely represent the signal of interest. Second, it estimates the parameters of the prior distribution from the sparse codes of the training data. Therefore, unlike most of the works that use overcomplete learned dictionaries in CS problems~\cite{Bilg12, Mair12, Zhou12}, which only use the training stage to learn the dictionary and disregard the sparse codes associated with such a dictionary, the proposed approach exploits both dictionary and sparse codes from the training stage to improve CS reconstruction algorithm performance.

In addition to the training stage, another contribution of the proposed approach is related to the reconstruction algorithm. After either RBMs or DBNs are employed to model the prior distribution of the sparsity pattern of the signal to be recovered, the determined prior is then employed in a maximum \textit{a posteriori} approach for reconstruction. Obtaining the exact MAP estimator solution can become computational unfeasible since complexity increases exponentially with the signal length. To overcome this limitation, we propose a greedy approach realized by modifying the orthogonal matching pursuit--based algorithm proposed in~\cite{Pele12} to maximize the posterior distribution of the sparsity pattern. 

The motivation for using RBMs and DBNs is twofold. First, they possess tremendous representational power; second, inference and parameter learning can be efficiently achieved using contrastive divergence and greedy layer--wise training~\cite{Le08, Suts08, Beng09}. Indeed, Le Roux \textit{et al}.~\cite{Le08} showed that an RBM can model any discrete distribution. Moreover, adding hidden units yields strictly enhanced modeling performance, unless the RBM already perfectly models the data. Similarly, Sutskever \textit{et al}.~\cite{Suts08} showed that deep belief networks can approximate any distribution over binary vectors to an arbitrary level of accuracy, even when the width of each layer is limited to the dimensionality of the data. Deep belief networks is one of the architectures of deep learning, a powerful and fast--growing field in artificial intelligence~\cite{Beng09}. Therefore, this manuscript links deep learning with CS by exploring the capabilities of deep learning architectures in modeling the statistical dependencies in the sparsity pattern of signals. To the best of our knowledge, this is the first paper that uses deep learning--based priors to model the sparsity pattern of signals in a compressed sensing framework. 

Previous works have employed fully visible Boltzmann machines to model the signal support in the context of compressed sensing~\cite{Cevh09, Pele12} and sparse coding~\cite{Garri08, Drem12}. Restricted Boltzmann machines have been employed to model the dependencies between low resolution and high resolution patches in the image super--resolution problem~\cite{Pele14}. The work of Tramel \textit{et al}.~\cite{Trame15} also uses RBMs to model the sparsity pattern of signals. Their work is based on the approximate message passing (AMP) framework, which is very powerful at reconstructing sparse signals by exploiting the statistical properties of the problem. However, it has been shown that AMP algorithms are very sensitive to parameter tuning~\cite{Male10}. Regardless of their simplicity and ease of implementation, OMP-based algorithms outperform AMP algorithms in some cases~\cite{Naka16, Zhen16}, specially when the non-zero coefficients of sparse signals differ in magnitude~\cite{Zhen16}.

The closest related work to ours was proposed by Peleg \textit{et al}.~\cite{Pele12}. They used fully visible Boltzmann machines to model the distribution of the sparsity pattern of sparse signals. We follow their method in trying to reconstruct the signal support using a MAP approach. Our work differs from that of Peleg \textit{et al}. in several aspects. First, the aim of their work is to learn sparse representations for signal modeling. The aim of our work is different, namely to reconstruct sparse signals from undersampled measurements. Second, they use fully visible Boltzmann machines that can only model pair--wise dependencies between elements in the sparsity pattern. Instead, RBMs and DBNs can model higher--order dependencies and, therefore, they offer superior representational power. Third, we employ contrastive divergence for parameter learning instead of the maximum  pseudo--likelihood approach in order to realize computational complexity and performance improvements. In practice, pseudo--likelihood learning has a high computational overhead compared to contrastive divergence~\cite{Laro12}. Pseudo--likelihood learning does not approximate the maximum likelihood estimator well, except in the limit of zero dependence~\cite{Geye91}. It was shown that contrastive divergence is equivalent to pseudo--likelihood for fully visible Boltzmann machines if single--step Gibbs sampling is employed and outperforms pseudo--likelihood when the number of sampling steps is larger than one~\cite{Asun10}. 

The organization of the paper is as follows. Section II presents a brief review of CS and deep learning architectures. In Section III, the proposed method is presented. Numerical results for the proposed method and comparisons with CS reconstruction algorithms are presented in Section IV. Finally, Section V concludes the work with closing remarks.

%Note that
%the sampling is not adaptive, is just optimized for the signal
%class.
%
%They also showed that adding hidden layers always increases the representational power of the DBN unless it is already exponentially deep.\mathbf{S
%
%The learned dictionary and sensing matrix do not need to be
%ommunicated to the reconstruction side at the rate of one per
%mage (in case of the use of this framework for transmission),
%nce these are learned for image groups or classes. In the case
%f the examples presented in this paper, these are learned for the
%arge class of natural images from a standard dataset.

\section{Background}
\subsection{Compressed sensing}
Let $\mathbf{x}\in\mathbb{R}^N$ be a signal that is approximately $K$-sparse in a dictionary $\mathbf{D}\in\mathbb{R}^{N\times Q}$. Thus, the signal $\mathbf{x}$ can be approximated by a linear combination of a small number of column vectors from $\mathbf{D}$, \textit{i.e.} $\mathbf{x}=\mathbf{Ds}+\mathbf{r}$, where $\mathbf{s}$ is the sparse vector of weighting coefficients, $\mathbf{r}$ is the representation error, and $K\ll N$. The support of $\mathbf{s}$ is denoted as $\theta$ and is associated with the sparsity pattern $\mathbf{S}$, which is defined as ${S}_i=\textbf{1}_{\theta}(i)$ for $i=1, \ldots, N$, where $\textbf{1}_{\theta}(i)$ takes the value of one if $i\in\theta$ and zero elsewhere. 
%$\mathbf{x}\approx\sum_{i=1}^{K} {s}_i\mathbf{D}_{\cdot i}$

Let $\mathbf{\Phi}$ be an $M\times N$ sensing matrix, $M<N$. Compressed sensing~\cite{Dono06, Cand08} addresses the recovery of $\mathbf{x}$ from linear measurements of the form $\mathbf{y}=\mathbf{\Phi x}\approx \mathbf{\Phi D s}$. Compressed sensing results show that the signal $\textbf{x}$ can be reconstructed from $\textbf{y}$ if the matrix $\mathbf{\Xi}=\mathbf{\Phi D}$ satisfies a condition, known as the restricted isometry property (RIP)~\cite{Cand06b}, with a sufficiently small restricted isometry constant. If the sampling matrix $\mathbf{\Phi}$ is a sub--Gaussian matrix and the dictionary $\mathbf{D}$ is unitary, then the matrix $\mathbf{\Xi}$ satisfies the RIP with high probability. However, if the matrix $\mathbf{D}$ is overcomplete, the coherence between its columns makes it difficult for the matrix $\mathbf{\Xi}$ to satisfy the RIP with a sufficiently small restricted isometry constant~\cite{Rauh08}. Several works have addressed this limitation by designing the sampling matrix $\mathbf{\Xi}$ so as to minimize the mutual coherence of the effective dictionary $\mathbf{\Xi}$~\cite{Duar09, Elad07}.

\subsection{Deep learning architectures}
Deep learning aims at learning hierarchical feature representations with higher level features formed by the composition of lower level features~\cite{Beng09}. Deep learning is inspired by biological structures in human brain mechanisms for processing of natural signals. A deep learning architecture, the DBN~\cite{Hint06}, is presented in this section.

\begin{figure}[t]
\centering{ % \hfill
\includegraphics[width = 0.8\columnwidth]{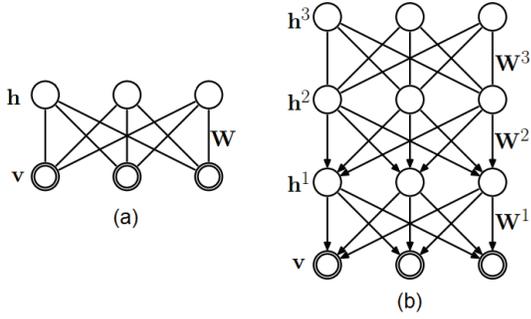}}
\caption{(a) Schematic of a restricted Boltzmann machine. (b) Schematic of a deep belief network of one visible and three hidden layers (adapted from~\cite{Sala09}).} \label{RBM}
\end{figure}

Restricted Boltzmann machines are the building blocks of DBNs. They are probabilistic generative models that learn a joint probability distribution of training data. An RBM is composed of a single visible layer and a single hidden layer. The visible units, $\mathbf{v}=[{v}_1 {v}_2 \ldots {v}_J]^T$, represent the input data whose probability distribution needs to be modeled. The hidden units, $\mathbf{h}=[{h}_1 {h}_2 \ldots {h}_P]^T$, are trained to capture higher-order data correlations that are observed at the visible units. Symmetric connections between the layers are represented by a weight matrix $\mathbf{W}$. The structure of an RBM forms a bipartite graph, as shown in Fig.~\ref{RBM}(a).

In an RBM, the joint distribution $p(\mathbf{v}, \mathbf{h})$, over the visible units $\mathbf{v}$ and the hidden units $\mathbf{h}$, is defined as
\begin{equation}\label{eq6.1}
p(\mathbf{v}, \mathbf{h})=\frac{\text{exp}(-\text{E}(\mathbf{v},\mathbf{h}))}{Z},
\end{equation}
where $\text{E}(\mathbf{v},\mathbf{h})$ is the energy function and $Z=\sum_\mathbf{v}\sum_\mathbf{h} \text{exp}(-\text{E}(\mathbf{v}, \mathbf{h}))$ is the normalization term. For a Bernoulli (visible)--Bernoulli (hidden) RBM, the energy function takes the form
\begin{equation}\label{eq6.2}
\text{E}(\mathbf{v}, \mathbf{h})=-\mathbf{v}^T\mathbf{Wh}-\mathbf{b}_{v}^T\mathbf{v}-\mathbf{b}_h^T\mathbf{h},
\end{equation}
where $\mathbf{W}$ denotes the weights between visible and hidden units, and $\mathbf{b}_v$ and $\mathbf{b}_h$ are the bias terms. The RBM parameters, \textit{i.e.}, $\mathbf{W}$, $\mathbf{b}_v$ and $\mathbf{b}_h$, can be optimized by performing stochastic gradient ascent on the log--likelihood of training data. Given that computing the exact gradient of the log--likelihood is intractable, the contrastive divergence approximation~\cite{Hint02} is typically employed. 

In an RBM, units within the same layer are not connected. Therefore, the posterior distribution over hidden vectors factorizes into a product of independent distributions for each hidden unit. The conditional distributions over hidden and visible units take the form
%\begin{equation}\label{eq6.4}
%({h}_j=1|\mathbf{v})=\sigma(({{b}_h})_j+\mathbf{W}_{\cdot j}^T\mathbf{v}),
%\end{equation}
%\begin{equation}\label{eq6.5}
%p(\mathbf{v}_i=1|\mathbf{h})=\sigma(({\mathbf{b}_v})_i+\mathbf{(W^T)}_{i}\mathbf{h}),
%\end{equation}

\begin{equation}\label{eq6.4}
p({h}_j=1|\mathbf{v})=\sigma(({{b}_h})_j+\mathbf{W}_{\cdot j}^{T}\mathbf{v}),
\end{equation}
\begin{equation}\label{eq6.5}
p({v}_i=1|\mathbf{h})=\sigma(({{b}_v})_i+\mathbf{W}_{i\cdot}^T\mathbf{h}),
\end{equation}

where $\sigma(x)=(1+e^{-x})^{-1}$, and $\mathbf{W}_{\cdot j}$ and $\mathbf{W}_{i\cdot}$ correspond to the $j$th column and $i$th row of matrix $\mathbf{W}$, respectively. 

A DBN architecture is composed of a stack of RBMs. The lowest--level RBM learns a shallow model of the data. The RBM at the next level captures high--order correlations between the hidden units of the first, and so on. A DBN with $L$ layers models the joint distribution between the visible layer $\mathbf{v}$ and the hidden layers $\mathbf{h}^l$, $l=1, \ldots, L$ as follows
\begin{equation}\label{eq6.3}
p(\mathbf{v}, \mathbf{h}^1, \ldots, \mathbf{h}^L)=p(\mathbf{v}|\mathbf{h}^{1})\left(\prod_{l=1}^{L-2}p(\mathbf{h}^l|\mathbf{h}^{l+1})\right)p(\mathbf{h}^{L-1},\mathbf{h}^L).
\end{equation}
The log-probability of the training data can be improved by adding layers to the network, which, in turn, increases the true representational power of the network~\cite{Hint06}.

Let $\mathbf{v}=\mathbf{h}^0$. The bias vector of layer $l$ and the weight matrix that represents the connections between layer $l-1$ and layer $l$ are denoted by $\mathbf{b}_{\mathbf{h}^l}$ and $\mathbf{W}^l$, respectively. A schematic representation of a DBN with one visible and three hidden layers is shown in Fig.~\ref{RBM}(b). The top two layers form a restricted Boltzmann machine,  which is an undirected graphical model, and the lower layers form a directed generative model.

The main breakthrough introduced by Hinton \textit{et al}.~\cite{Hint06} was a greedy, layer--wise unsupervised learning algorithm that allows efficient training of DBNs. This algorithm trains each RBM separately, making the time complexity of the DBN learning linear in the size and depth of the networks.

\section{Proposed Compressed Sensing Scheme}
The proposed scheme requires training data of the same class as the signal to be reconstructed. A training stage is employed to learn a prior model for the sparsity pattern of the signal class. The proposed CS reconstruction algorithm exploits the determined prior in a MAP approach. The CS and training stages are described thoroughly in this section. The training stage varies depending on the employed sparsifying transform, either orthonormal bases or overcomplete learned dictionaries. The block diagrams of the proposed CS schemes for overcomplete dictionaries and orthonormal bases are presented in Figs.~\ref{block2} and~\ref{block1}, respectively.

\subsection{Compressed sensing stage}
\subsubsection{Problem formulation}
\label{sec:pf}
Let $\mathbf{D}\in\mathbb{R}^{N\times Q}$ denote the sparsifying transform employed to represent a signal $\mathbf{x}\in\mathbb{R}^N$, \textit{i.e.}, $\mathbf{x}=\mathbf{Ds}+\mathbf{r}$, where $\mathbf{s}$ and $\mathbf{r}$ are the sparse representation and the representation error, respectively. A Gaussian distribution with zero mean and covariance $\mathbf{\Sigma}_\mathbf{r}$ is assumed for $\mathbf{r}$. In this paper, we consider the traditional synthesis--based CS approach that aims at reconstructing the sparse representation $\mathbf{s}$ of a signal $\mathbf{x}$ from undersampled and noisy measurements of the form $\mathbf{y}=\mathbf{\Phi x}+\mathbf{n}$, where $\mathbf{\Phi} \in\mathbb{R}^{M\times N}$ is the sampling matrix and $\mathbf{n}$ accounts for the additive Gaussian sampling noise of zero mean and variance $\sigma_n^2$. Vector $\mathbf{y}$ can also be written as
\begin{equation}\label{eq6.24}
\mathbf{y}=\mathbf{\Phi Ds}+\mathbf{\Phi r}+\mathbf{n}.
\end{equation}

Let $\bm{\eta}=\mathbf{\Phi r}+\mathbf{n}$ and $\mathbf{\Xi}=\mathbf{\Phi D}$, then vector $\mathbf{y}$ takes the form
\begin{equation}\label{ojala}
\mathbf{y}=\mathbf{\Xi s}+\bm{\eta}.
\end{equation}
As both $\mathbf{r}$ and $\mathbf{n}$ are Gaussian distributed, vector $\bm{\eta}$ is also Gaussian distributed with zero mean and covariance $\mathbf{\Sigma}_\mathbf{\eta}=\mathbf{\Phi\Sigma}_\mathbf{r}\mathbf{\Phi}^T+\sigma_n^2\mathbf{I}$.  We adopt the commonly used assumption that the sampling noise variance $\sigma_n^2$ is known~\cite{Yang13, Pele12, Pras13}.

\begin{figure}[t]
\centering{ % \hfill
\includegraphics[width = \columnwidth]{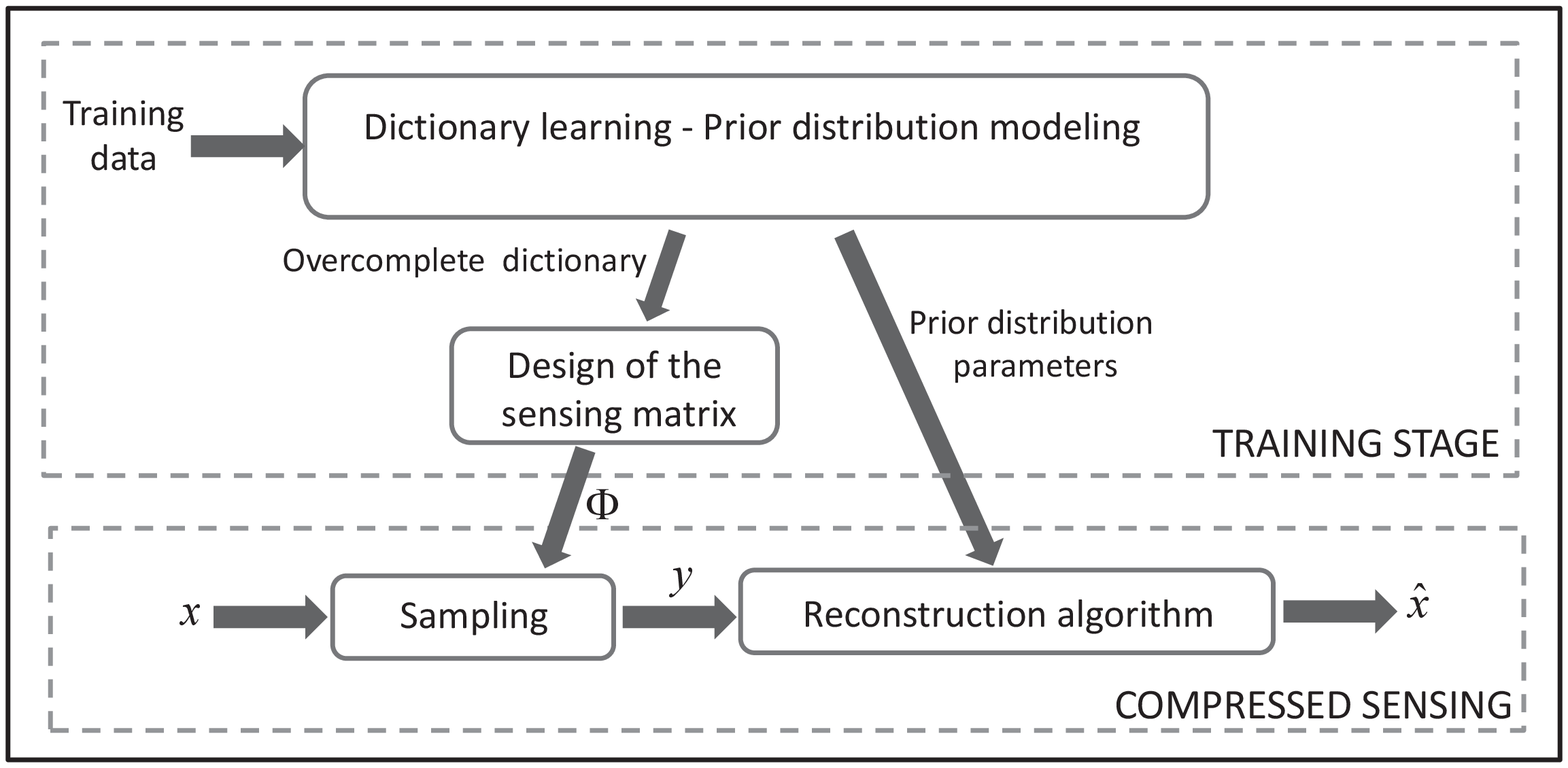}}
\caption{Block diagram of the proposed CS scheme when using overcomplete learned dictionaries as the sparsifying transform.} \label{block2}
\end{figure}

\begin{figure}[t]
\vspace{-.3cm}
\centering{ % \hfill
\includegraphics[width = \columnwidth]{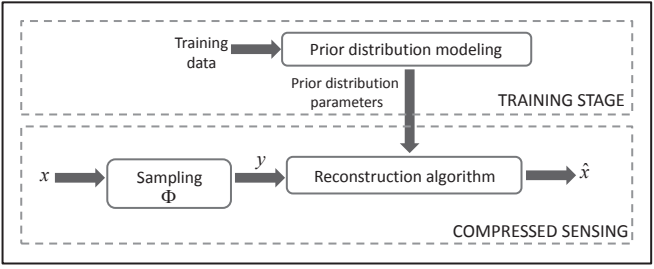}}
\caption{Block diagram of the proposed CS scheme when using orthonormal bases as the sparsifying transform.} \label{block1}
\end{figure}

The approach proposed in~\cite{Pele12} is adopted in this paper, namely first calculating the MAP estimator of the sparsity pattern and then calculating the MAP estimator of the sparse vector. The support of $\mathbf{s}$, of cardinality $K$, is denoted as $\theta$. Let $\mathbf{s}_\theta$ denote the nonzero coefficients of $\mathbf{s}$. A Gaussian distribution with zero mean and variance $\sigma_{{s}_i}^2$ is assumed for each nonzero coefficient ${s}_i$, $i\in \theta$. The same probability distribution is employed in~\cite{Pele12, Garr08} for nonzero sparse coefficients. Then, the conditional distribution of $\mathbf{s}_\theta$ given $\theta$ is given by
\begin{equation}\label{eq6.8}
\mathbf{s}_\theta|\theta \sim \mathcal{N}(\mathbf{0}, \mathbf{\Sigma}_\theta),
\end{equation}
where $\mathbf{\Sigma}_\theta\in\mathbb{R}^{K\times K}$ is a diagonal matrix, whose diagonal is formed by the variances of the nonzero coefficients $\sigma_{s_i}^2$, $i\in \theta$.

The Gaussian distribution of $\bm{\eta}$ leads to the following distribution for the likelihood $\text{p}(\mathbf{y}|\mathbf{s}_\theta, \theta)$:
\begin{equation}\label{eq6.9}
\mathbf{y}|\mathbf{s}_\theta, \theta \sim \mathcal{N}(\mathbf{\Xi}_\theta \mathbf{s}_\theta, \mathbf{\Sigma}_\mathbf{\eta}),
\end{equation}
where $\mathbf{\Xi}_\theta$ is the submatrix obtained by extracting the columns of matrix $\mathbf{\Xi}$ corresponding to the indexes in $\theta$. By integrating the product of $\text{p}(\mathbf{s}_\theta|\theta)$ and $\text{p}(\mathbf{y}|\mathbf{s}_\theta, \theta)$ over all possible $\mathbf{s}_\theta$, an expression for the probability distribution $\text{p}(\mathbf{y}|\theta)$ is obtained,
\begin{equation}\label{eq6.10}
\begin{split}
\text{p}(\mathbf{y}|\theta)=&C\times{\text{det}\left(\mathbf{\Xi}_\theta^T\mathbf{\Sigma}_\mathbf{\eta}^{-1}\mathbf{\Xi}_\theta\mathbf{\Sigma}_\theta+\mathbf{I}\right)^{-1/2}}\\
   &\times\text{exp}\left\{\frac{1}{2}\mathbf{y}^T\mathbf{\Sigma}_\mathbf{\eta}^{-1}\mathbf{\Xi}_\theta \mathbf{P}^{-1}\mathbf{\Xi}_\theta^T\mathbf{\Sigma}_\mathbf{\eta}^{-1}\mathbf{y}\right\},
\end{split}
\end{equation}
where $C={{\text{det}(2\pi\mathbf{\Sigma}_\mathbf{\eta})}^{-1/2}}\text{exp}\left\{-\frac{1}{2}\mathbf{y}^T\mathbf{\Sigma}_\mathbf{\eta}^{-1}\mathbf{y}\right\}$ and $\mathbf{P}=\mathbf{\Xi}_\theta^T\mathbf{\Sigma}_\mathbf{\eta}^{-1}\mathbf{\Xi}_\theta+\mathbf{\Sigma}_\theta^{-1}$.

The MAP estimator of $\theta$, denoted by $\hat{\theta}$, can be calculated as
\begin{equation}\label{eq6.111}
\hat{\theta}=\argmax_{\theta} p(\theta|\mathbf{y})=\argmax_{\theta} p(\mathbf{y}|\theta)p(\theta).
\end{equation}
The posterior distribution $p(\mathbf{s}_{\hat{\theta}}|\mathbf{y}, \hat{\theta})$ has a Gaussian distribution with mean $\bm{\mu}_\mathbf{s}$ and covariance $\mathbf{\Sigma}_\mathbf{s}$, such that
\begin{eqnarray}
\bm{\mu}_\mathbf{s}&=&\mathbf{\Sigma}_{\hat{\theta}}\mathbf{\Xi}_{\hat{\theta}}^T(\mathbf{\Xi}_{\hat{\theta}}\mathbf{\Sigma}_{\hat{\theta}} \mathbf{\Xi}_{\hat{\theta}}^T+\mathbf{\Sigma}_\mathbf{\eta})^{-1}\mathbf{y}\\
\mathbf{\Sigma}_\mathbf{s}&=&\mathbf{\Sigma}_{\hat{\theta}}-\mathbf{\Sigma}_{\hat{\theta}} \mathbf{\Xi}_{\hat{\theta}}^T(\mathbf{\Xi}_{\hat{\theta}}\mathbf{\Sigma}_{\hat{\theta}}\mathbf{\Xi}_{\hat{\theta}}^T+\mathbf{\Sigma}_\eta)^{-1}\mathbf{\Xi}_{\hat{\theta}}\mathbf{\Sigma}_{\hat{\theta}}.
\end{eqnarray}
Therefore, the MAP estimate of $\mathbf{s}$, denoted as $\hat{\mathbf{s}}_{\hat{\theta}}$, is directly obtained from the mean of the posterior, \textit{i.e.},
\begin{eqnarray}\label{eq6.18}
\hat{\mathbf{s}}_{\hat{\theta}}&=&\argmax_{\mathbf{s}_{\hat{\theta}}} p(\mathbf{s}_{\hat{\theta}}|\mathbf{y}, \hat{\theta}),\\
&=&\mathbf{\Sigma}_{\hat{\theta}} \mathbf{\Xi}_{\hat{\theta}}^T(\mathbf{\Xi}_{\hat{\theta}}\mathbf{\Sigma}_{\hat{\theta}} \mathbf{\Xi}_{\hat{\theta}}^T+\mathbf{\Sigma}_\mathbf{\eta})^{-1}\mathbf{y}. \nonumber
\end{eqnarray}

An expression for $p(\theta)$ needs to be calculated to solve \eqref{eq6.111}. Note that $p(\theta)=p\left(\mathbf{S}\right)$. We propose to use the probability distribution over visible units $p(\mathbf{v})$ of RBMs and DBNs to model the prior distribution $p\left(\mathbf{S}\right)$, or equivalently, $p(\theta)$. 

Using RBMs to model the sparsity pattern probability distribution is justified by results that show that an RBM can model any discrete distribution and that adding hidden units yields strictly enhanced modeling performance, unless the RBM already perfectly models the data~\cite{Le08}. Similarly, the use of deep belief networks is justified by their capabilities to approximate any distribution over binary vectors to an arbitrary level of accuracy, even when the width of each layer is limited to the dimensionality of the data~\cite{Suts08}. However, DBNs offer an additional advantage over RBMs: they yield more efficient and compact representations in terms of the number of parameters~\cite{Le08}.

\subsubsection{Prior distribution}
In an RBM, the probability distribution over visible units is obtained by marginalizing \eqref{eq6.1} over the hidden units
\begin{equation}\label{eq6.12}
p(\mathbf{v})=\sum_{\mathbf{h}} p(\mathbf{v},\mathbf{h})=-\frac{1}{Z}\text{exp}\left(-\text{E}(\mathbf{v})\right),
\end{equation}
where
\begin{equation}\label{eq6.13}
\text{E}(\mathbf{v})=-\sum_j\text{log}\left(1+e^{\mathbf{W}_{\cdot j}^T\mathbf{v}+{{b}_{h}}_j}\right)-\mathbf{b}_\mathbf{v}^T\mathbf{v}.
\end{equation}

A DBN can be seen as a probabilistic generative model. To calculate the probability distribution of the visible units of a DBN, we start with a random configuration at the top hidden layer, $\mathbf{h}^L$, and let the top--level RBM converge to a stationary distribution using alternating Gibbs sampling. Alternating Gibbs sampling iterates between updating the hidden units in parallel using \eqref{eq6.4}  and updating the visible units in parallel using \eqref{eq6.5}. Next, it performs a top--down pass in which the state of each variable in a layer is chosen from a Bernoulli distribution with the probability that a variable has a value of one depending on the states of the layer above. That is, \begin{equation}\label{eq6.14}
p({h}^l_i=1|\mathbf{h}^{l+1})=\sigma(\left({{b}_{{h}^l}}\right)_i+\mathbf{W}_{i\cdot}^{l+1}\mathbf{h}^{l+1}),
\end{equation}
where, as before, $\mathbf{v}=\mathbf{h}^0$.

Repeated top--down passes generates a full set of data vectors at each layer of the DBN. Let $H$ be the total number of top--down passes. The sequence of data vectors for the hidden layers is denoted as ${\mathbf{h}^l}^{(1)}, \ldots, {\mathbf{h}^l}^{(H)}$, $l=1,...,L$, and $\mathbf{v}^{(1)}, \ldots, \mathbf{v}^{(H)}$, for the visible layer.  Such sequence of data vectors assigned to the visible layer can be employed to give a rough approximation of the marginal distribution $p(\mathbf{v})$. However, the conditional density function $p(\mathbf{v}|\mathbf{h}^1)$ contains more information about the shape of the distribution $p(\mathbf{v})$ than the sequence of the individual realizations since $p(\mathbf{v})=E_{\mathbf{h}^1}[p(\mathbf{v}|\mathbf{h}^1)]$, where $E[\cdot]$ denotes the expected value of a random variable. Therefore, the marginal density is approximated by $\hat{p}(\mathbf{v})=\frac{1}{H}\sum_{k=1}^H p\left(\mathbf{v}\mathbf{|}{\mathbf{h}^1}^{(k)}\right)$. Since the visible units are conditionally independent given the hidden states $\mathbf{h}^1$, the approximation of $p(\mathbf{v})$ takes the form
\begin{equation}\label{eq6.15}
\hat{p}(\mathbf{v})=\frac{1}{H}\sum_{k=1}^H\prod_{i=1}^N p\left({v}_i|{\mathbf{h}^1}^{(k)}\right).
\end{equation}

Using the probability distribution \eqref{eq6.12} to model $p(\theta)$ in \eqref{eq6.111} leads to the following MAP estimator of $\theta$:

\begin{eqnarray}\label{eq6.16}
\hat{\theta}&=&\argmax_{\theta} \left(\frac{1}{2}\mathbf{y}^T\mathbf{\Sigma}_\mathbf{\eta}^{-1}\mathbf{\Xi}_\theta \mathbf{P}^{-1}\mathbf{\Xi}_\theta^T\mathbf{\Sigma}_\eta^{-1}\mathbf{y}\right.\\
&&-\frac{1}{2}\text{log}\left(\text{det}(\mathbf{P}\mathbf{\Sigma}_\theta)\right)+\sum_j\text{log}\left(1+e^{\mathbf{W}_{\cdot j}^T\mathbf{\mathbf{S}}+{{b}_{{h}}}_j}\right) \nonumber\\ 
&&\left.+\mathbf{b}_\mathbf{v}^T\mathbf{S}\right ). \nonumber
\end{eqnarray}

%\begin{multline}\label{eq6.16}
%\hat{\theta}=\argmax_{\theta} \left(\frac{1}{2}\mathbf{y}^T\mathbf{\Sigma}_\mathbf{\eta}^{-1}\mathbf{\Xi}_\theta \mathbf{P}^{-1}\mathbf{\Xi}_\theta^T\mathbf{\Sigma}_\eta^{-1}\mathbf{y}-\frac{1}{2}\text{log}\left(\text{det}(\mathbf{P}\mathbf{\Sigma}_\theta)\right)\right.\\ \left.+\sum_j\text{log}\left(1+e^{\mathbf{W}_{\cdot j}^T\mathbf{\mathbf{S}}+{{b}_{{h}}}_j}\right)
%+\mathbf{b}_\mathbf{v}^T\mathbf{S}\right).
%\end{multline}

Similarly, when using \eqref{eq6.15} to model $p(\theta)$ in \eqref{eq6.111}, the MAP estimator of $\theta$ becomes

\begin{eqnarray}\label{eq6.17}
\hat{\theta}&=&\argmax_{\theta} \left (\frac{1}{2}\mathbf{y}^T\mathbf{\Sigma}_\mathbf{\eta}^{-1}\mathbf{\Xi}_\theta \mathbf{P}^{-1}\mathbf{\Xi}_\theta^T\mathbf{\Sigma}_\eta^{-1}\mathbf{y}\right.\\
&&-\frac{1}{2}\text{log}\left(\text{det}(\mathbf{P}\mathbf{\Sigma}_\theta)\right)\left.+\text{log}\left(\sum_{k=1}^H\prod_{i=1}^N p\left({S}_i|{\mathbf{h}^1}^{(k)}\right)\right)\right) \nonumber
\end{eqnarray}

%\begin{multline}\label{eq6.17}
%\hat{\theta}=\argmax_{\theta} \left(\frac{1}{2}\mathbf{y}^T\mathbf{\Sigma}_\mathbf{\eta}^{-1}\mathbf{\Xi}_\theta \mathbf{P}^{-1}\mathbf{\Xi}_\theta^T\mathbf{\Sigma}_\eta^{-1}\mathbf{y}-\frac{1}{2}\text{log}\left(\text{det}(\mathbf{P}\mathbf{\Sigma}_\theta)\right)\right.\\
%\left.+\text{log}\left(\sum_{k=1}^H\prod_{i=1}^N p\left({S}_i|{\mathbf{h}^1}^{(k)}\right)\right)\right)
%\end{multline}

%\begin{multline}\label{eq6.17}
%\hat{\theta}=\argmax_{\theta} \left(\frac{1}{2}\mathbf{y}^T\mathbf{\Sigma}_\mathbf{\eta}^{-1}\mathbf{\Xi}_\theta \mathbf{P}^{-1}\mathbf{\Xi}_\theta^T\mathbf{\Sigma}_\eta^{-1}\mathbf{y}-\frac{1}{2}\text{log}\left(\text{det}(\mathbf{P}\mathbf{\Sigma}_\theta)\right)\right.\\
%\left.+\text{log}\left(\sum_{k=1}^H\prod_{i=1}^N p\left({S}_i|{\mathbf{h}^1}^{(k)}\right)\right)\right)
%\end{multline}

Expressions \eqref{eq6.16} and \eqref{eq6.17} are combinatorial optimization problems that require an exhaustive search over all possible sparsity patterns of dimension $K$. Therefore, there is a need to use algorithms that approximate the MAP solution to overcome the intractability of the combinatorial search. 

\subsubsection{MAP estimator via a greedy approach}
\label{sec:greedy}
Peleg \textit{et al}.~\cite{Pele12} proposed a greedy pursuit algorithm based on Orthogonal Matching Pursuit (OMP) to approximate the MAP estimator of a sparse representation. The same algorithm is used here for reconstructing the sparse representation $s$, although using a different posterior distribution than that in~\cite{Pele12}. The support is initialized to the empty set. At each iteration, the algorithm searches for the element $\bar{i}$ that can be added to the support in order to maximize $p(\theta|\mathbf{y})$. The algorithm stops when any additional element in the support decreases either the objective function in \eqref{eq6.16}, in the case of RBMs, or \eqref{eq6.17}, in the case of DBNs. The algorithm, at each iteration, makes locally optimal choices with the hope that this will lead to the optimal global solution. It is noted that the algorithm achieves the optimal solution for supports of cardinality 1 since it goes through the same computational stages as exhaustive search in this special case. 

Once the support is recovered, the sparse representation $\mathbf{s}$ is calculated using~\eqref{eq6.18}. A summary of the algorithm is presented in Algorithm~\ref{alg:alg1}. The functions $g_{\text{RBM}}(\cdot)$ and $g_{\text{DBN}}(\cdot)$ refer to the objective functions in~\eqref{eq6.16} and ~\eqref{eq6.17}, respectively. If an RBM is employed to model the prior distribution $p(\theta)$, the function $g_{\text{RBM}}(\cdot)$ is used and the algorithm is referred to as the RBM--OMP--like algorithm. If a DBN is employed instead, the function $g_{\text{DBN}}(\cdot)$ is used and the algorithm is referred to as the DBN--OMP--like algorithm.

\begin{algorithm}[t]
\caption{RBM--OMP--like/DBN--OMP--like algorithm}\label{alg:alg1}
\begin{algorithmic}[1]
\REQUIRE Matrix $\mathbf{\Xi}=\mathbf{\Phi D}$, measurements $\mathbf{y}$, model parameters defining the probability distribution $p(\theta|\mathbf{y})$.
\STATE Initialize $t=0$, ${\theta}^{(0)}=\emptyset$.
\WHILE{halting criterion false}
\STATE $t\leftarrow t+1$
\FOR{$i\notin {\theta}^{(t-1)}$}
\STATE $\bar{\theta}^{(t)}\leftarrow {\theta}^{(t-1)}\cup i$
\STATE $f(i)\leftarrow g_{\text{RBM}}(\bar{\theta}^{(t)})$ or $f(i)\leftarrow g_{\text{DBN}}(\bar{\theta}^{(t)})$
\ENDFOR
\STATE $\bar{i}\leftarrow %\displaystyle
\argmax_{i} f(i)$
%\vspace{.2cm}
\STATE ${\theta}^{(t)}\leftarrow {\theta}^{(t-1)}\cup \bar{i}$
\ENDWHILE
\STATE $\hat{\theta} \leftarrow {\theta}^{(t)}$
\RETURN  $\hat{\mathbf{s}}_{\hat{\theta}}\leftarrow\mathbf{\Sigma}_{\hat{\theta}} \mathbf{\Xi}_{\hat{\theta}}^T(\mathbf{\Xi}_{\hat{\theta}}\mathbf{\Sigma}_{\hat{\theta}} \mathbf{\Xi}_{\hat{\theta}}^T+\mathbf{\Sigma}_\mathbf{\eta})^{-1}\mathbf{y}$.
\end{algorithmic}
\end{algorithm}

\subsubsection{Computational Complexity Analysis}
\label{sec:complexity}
The computational complexity in Algorithm 1 is dominated by the calculation of the functions in lines 6, $g_{\text{RBM}}(\bar{\theta}^{(t)})$ and $g_{\text{DBN}}(\bar{\theta}^{(t)})$. It is assumed that all the operations that do not depend on $\bar{\theta}^{(t)}$ are precomputed and do not contribute to the cost. Evaluation of the first two terms of $g_{\text{RBM}}(\bar{\theta}^{(t)})$ is dominated by the calculation of $\mathbf{P}$, which costs $\mathcal{O}(KM^2)$ flops. The third and fourth terms cost $\mathcal{O}(NP)$ and $\mathcal{O}(N)$ flops, respectively. Therefore, calculating $g_{\text{RBM}}(\cdot)$ costs $\mathcal{O}(KM^2+NP)$ flops. Let $D$ denote the total number of iterations of the while loop in Algorithm 1. Since $g_{\text{RBM}(\cdot)}$ needs to be calculated $D(N-1)$ times, the computational complexity of Algorithm~1 is $\mathcal{O}(DN(KM^2+NP))$, when RBMs are employed.

The first two terms of $g_{\text{DBN}}(\bar{\theta}^{(t)})$ are the same as those of $g_{\text{RBM}}(\bar{\theta}^{(t)})$, which means that their cost is $\mathcal{O}(KM^2)$ flops. Let $C$ denote the number of units of the first hidden layer of the DBN. Then, calculating the third term of $g_{\text{DBN}}(\bar{\theta}^{(t)})$ costs $\mathcal{O}(CHN)$ flops and, therefore, the computational complexity of Algorithm~1 becomes $\mathcal{O}(DN(KM^2+CNH))$, when DBNs are employed.

%The calculation of $\hat{\mathbf{s}}_{\hat{\theta}}$ in line 12 costs $\mathcal{O}(M^3)$ flops due to the matrix inversion. 
\subsubsection{Selection of the probabilistic generative model}
\label{sec:selmodel}
Even though DBNs and RBMs have the same representational power~\cite{Le08}, they differ in the number of parameters needed to model the probability distribution of the data. Deep Belief Networks offer more compact representations. Indeed, in~\cite{Beng09}, it was shown that deep architectures can sometimes be exponentially more efficient than shallow ones in terms of number of parameters needed to represent a function. More precisely, there are functions that can be compactly represented with an architecture of depth $L$ but that would require exponential size (with respect to input size) architectures of depth $L-1$.

The number of parameters have computational and statistical consequences, and, therefore, can be used as a factor to decide which generative model to use. Since the number of parameters one can afford depends on the number of training samples available to learn them, the selection of RBMs, which typically require more parameters than DBNs, may lead to poor generalization. Also, the computational cost of the training stage and reconstruction algorithms increases with the number of parameters. 

This result may initially suggest that DBNs should always be selected. However, as shown in Section \ref{sec:complexity}, the computational complexity of the DBN-OMP-like algorithm also depends on the number of top-down passes used for the approximation of the probability distribution of the visible layer. Therefore, the decision of which generative model to use is data dependent as the exact number of required parameters depends on the complexity of the data and the amount of training data. We suggest to estimate the number of parameters of the generative models using cross-validation, as described in Section \ref{sec:num_hidden}, and then choose the generative model that would lead to the lowest computational cost of the reconstruction algorithm using the results in Section \ref{sec:complexity}.

\subsection{Training stage using Overcomplete Learned Dictionaries}
\label{sssec:DLP}
In this section,  the sparsifying transform $\mathbf{D}$ from \eqref{eq6.24} is assumed to be an overcomplete dictionary. We learn $\mathbf{D}$ from a set of training data belonging to the same class as signal $\mathbf{x}$. The resulting sparse codes and the representation error are employed to estimate the model parameters defining $p(\theta|\mathbf{y})$.

\subsubsection{Traditional dictionary learning}
\label{sssec:dictionary}
 First, the dictionary learning problem is described. Let $\mathbf{G}=[\mathbf{G}_{\cdot 1} \ldots \mathbf{G}_{\cdot B}] \in \mathbb{R}^{N\times B}$ denotes the set of $N$-dimensional training samples, which is referred to as training dataset I. One methodology for building the overcomplete dictionary $\mathbf{D}=[\mathbf{D}_{\cdot 1} \ldots \mathbf{D}_{\cdot J}] \in \mathbb{R}^{N\times J}$ $\left(J>N\right)$ is to solve the following optimization problem
\begin{equation}\label{eq6.6}
\{\hat{\mathbf{D}}, \hat{\mathbf{A}}\}=\argmin_{\mathbf{D}, \mathbf{A}} \|\mathbf{G}-\mathbf{DA}\|_{F}^2~~\textrm{s.~t.}~~\|\mathbf{A}_{\cdot j}\|_{0}<K,~~\forall j
\end{equation}
with $\mathbf{A}=[\mathbf{A}_{\cdot 1} \ldots \mathbf{A}_{\cdot B}] \in \mathbb{R}^{J\times B}$ denoting the sparse codes of $\mathbf{G}$, and $K$ the pre-specified sparsity level. The representation error is defined as $\mathbf{E}=\mathbf{G}-\hat{\mathbf{D}}\hat{\mathbf{A}}$. Several algorithms have been proposed to solve \eqref{eq6.6}. Here, the K--SVD algorithm proposed by Aharon \textit{et al.}~\cite{Ahar06} is employed. Rubinstein \textit{et al.}~\cite{rubi08} show that the dominant operations in K-SVD are sparse-coding, atom updates and coefficients updates, wich leads to a total computational complexity of $\mathcal{O}(B(K^3+2KNJ+4NK+4KJ)+5NJ^2)$.

\subsubsection{Sampling matrix design}
\label{sssec:matrix}
Even though an overcomplete dictionary offers more flexibility and leads to sparser representations compared to an orthonormal basis, the coherence between its columns limits the performance of CS recovery algorithms when using the traditional sub--Gaussian sensing matrix~\cite{Rauh08, Elad07, Duar09}. That is why several works have focused on how to design the sampling matrix to guarantee accurate recoverability when using an overcomplete dictionary as the sparsifying transform. For example, Elad~\cite{Elad07} optimizes the sampling matrix to minimize the coherence of the effective dictionary. Similarly, Duarte \textit{et al}.~\cite{Duar09} also aims at minimizing the coherence of the effective dictionary, with their approach designed to make the columns of the effective dictionary as orthogonal as possible. In this paper, the method of Duarte \textit{et al}. is employed to build the sensing matrix as it has superior performance in terms of running time and accuracy of the reconstructed signal. That is, the sensing matrix $\mathbf{\Phi}$ is optimized as to minimize the mutual coherence of the effective dictionary $\mathbf{\Phi D}$.

\subsubsection{Estimation of model parameters}
\label{sssec:modeling}
This section addresses the parameter estimation of $p(\theta|\mathbf{y})$, which comprises the parameter estimation of $p(\theta)$, the estimation of the variances $\sigma_{{s}_i}^2$, $\forall i$, and the estimation of the covariance $\mathbf{\Sigma}_\mathbf{\eta}$. Let $\mathbf{U}_{\cdot j}$ denote the sparsity pattern of the sparse code $\mathbf{A}_{\cdot j}$, $j=1, \ldots, B$. The $i$th element of $\mathbf{U}_{\cdot j}$ is defined as ${U}_{ij}=\textbf{1}_{\text{supp}(\mathbf{A}_{
\cdot j})}(i)$, where $\text{supp}(\mathbf{A}_{\cdot j})$ denotes the support of $\mathbf{A}_{\cdot j}$. The set of vectors  $\mathbf{U}=[\mathbf{U}_{\cdot 1} \ldots \mathbf{U}_{\cdot B}]$ can be employed to model a prior distribution for the sparsity pattern of signals belonging to the same class as the training data. Since the signal to be reconstructed, $\mathbf{x}$, belongs to this class, the set of column vectors of $\mathbf{U}$, which are referred to as training dataset II, are used to learn the parameters of $p(\theta)$. As mentioned in Section \ref{sec:pf}, RBMs and DBNs are used for modeling such a prior distribution.

In the case of the RBM model, the probability distribution parameters that need to be estimated are the weight matrix $\mathbf{W}$ and the bias terms $\mathbf{b}_\mathbf{v}$ and $\mathbf{b}_\mathbf{h}$. They can be optimized by performing stochastic gradient ascent on the log--likelihood of the training dataset II. However, computing the exact gradient of the log--likelihood is intractable. Here, we use contrastive divergence, which approximates the gradient of the log--likelihood by using a Markov chain. For more details on contrastive divergence, the reader is referred to~\cite{Hint02}.

When the prior distribution of the sparsity pattern is modeled with a DBN, the parameters that need to be estimated are the weights and bias terms of each layer. Hinton presented a powerful greedy layer--wise method to learn these parameters~\cite{Hint06}. The weight matrix $\mathbf{W}^1$, the bias terms of the visible layer $\mathbf{b}_\mathbf{v}$, and the bias terms of the lowest hidden layer $\mathbf{b}_{\mathbf{h}^1}$ are learned by training an RBM with the training dataset II as input. Then, the inferred hidden values of $\mathbf{h}^1$ can be used as input for training another RBM that learns the parameters of the layer above. This bottom--up process is repeated at the next layers until all the parameters of the network are learned. 

Additionally, the set of sparse codes $\mathbf{A}$ can be employed to estimate the variance $\sigma_{{s}_i}^2$ of each $i$th element of the sparse representation $\mathbf{s}$ of signal $\mathbf{x}$ (see ~\cite{Pele12}):
\begin{equation}\label{eq6.7}
\hat{\sigma}_{{s}_i}^2=\frac{\sum_{j=1}^B {A}_{ij}^2}{\sum_{j=1}^B \textbf{1}[i\in \text{supp}\left(\mathbf{A}_{\cdot j}\right)]}.
\end{equation}

For the estimation of the covariance matrix $\mathbf{\Sigma}_{\mathbf{r}}$, independence is assumed between the representation error coefficients ${r}_i$ and ${r}_j$ for $i\neq j$. Therefore, the covariance matrix $\mathbf{\Sigma}_\mathbf{r}$ is a diagonal matrix, whose diagonal is formed by the variances of the representation error coefficients $\sigma_{{r}_i}^2$, for $i=1, \ldots, N$. For the estimate of $\mathbf{\Sigma}_\mathbf{r}$, denoted as $\hat{\mathbf{\Sigma}}_\mathbf{r}$, the representation error of the learned dictionary $\mathbf{E}=[\mathbf{E}_{\cdot 1} \ldots \mathbf{E}_{\cdot B}]$ is employed. More precisely, each $i$th diagonal element of $\mathbf{\Sigma}_\mathbf{r}$ is estimated as
\begin{equation}\label{luisilla}
\hat{\sigma}_{{r}_i}^2=\frac{1}{B} \sum_{j=1}^B {E}_{ij}^2.
\end{equation}
The estimate of $\mathbf{\Sigma}_\mathbf{\eta}$ is directly calculated as $\hat{\mathbf{\Sigma}}_\mathbf{\eta}=\mathbf{\Phi}\hat{\mathbf{\Sigma}}_\mathbf{r}\mathbf{\Phi}^T+\sigma_n^2\mathbf{I}$.

\subsubsection{Selection of number of hidden layers and units}
\label{sec:num_hidden}
For the selection of the number of hidden layers and units of a DBN, hereafter referred to as hyperparameters, a validation dataset is employed. The selection of the hyperparameters is performed by grid search on the parameter space using holdout cross-validation. For every combination of hyperparameter values in the grid, the weights and biases of the DBN are learned using the training dataset I, which results in different DBN configurations. Those configurations are then used with the validation dataset. That is, the validation dataset is sampled with the sampling matrix $\mathbf{\Phi}$ and recovered with the DBN-OMP-like reconstruction algorithm using the different DBN configurations. The reconstruction SNR of the validation dataset, across the different architecture configurations, is used as a metric to determine the number of hidden layers and units. The same approach is employed for the selection of the number of hidden units in RBMs.

\subsection{Training stage using Orthonormal bases}
As in Section \ref{sssec:dictionary}, let $\mathbf{G}=[\mathbf{G}_{\cdot 1} \ldots \mathbf{G}_{\cdot B}] \in \mathbb{R}^{N\times B}$ denote the set of $N$-dimensional training samples belonging to the same class as signal $\mathbf{x}$. In this section, $\mathbf{D}$ does not denote an overcomplete dictionary, but instead, it denotes an orthonormal basis. Each vector $\mathbf{G}_{\cdot j}$ can be expressed as $\mathbf{G}_{\cdot j}=\mathbf{D}\bar{\mathbf{A}}_{\cdot j}$, where $\bar{\mathbf{A}}_{\cdot j}$ is the representation of the signal $\mathbf{G}_{\cdot j}$ in the $\mathbf{D}$ domain. Let $\mathbf{A}_{\cdot j}$ denote the best $K$-term approximation of $\bar{\mathbf{A}}_{\cdot j}$, which is obtained by keeping only the $K$ largest (in magnitude) coefficients in $\bar{\mathbf{A}}_{\cdot j}$ and setting the others to zero. Therefore, the signal $\mathbf{G}$ can be modeled as $\mathbf{G}=\mathbf{D}{\mathbf{A}}+{\mathbf{E}}$, where ${\mathbf{E}}$ is the representation error matrix. 

Let ${U}_{\cdot j}$ denote the sparsity pattern of the sparse code $\mathbf{A}_{\cdot j}$, $\mathbf{U}_{ij}=\textbf{1}_{\text{supp}(\mathbf{A}_{\cdot j})}(i)$. As in the case of overcomplete dictionaries, the sparse codes $\mathbf{A}=[\mathbf{A}_{\cdot 1} \ldots \mathbf{A}_{\cdot B}]$, $\mathbf{U}=[\mathbf{U}_{\cdot 1} \ldots \mathbf{U}_{\cdot B}]$, and the representation error $\mathbf{E}=[\mathbf{E}_{\cdot 1} \ldots \mathbf{E}_{\cdot B}]$ are used to learn the model parameters defining $p(\theta|\mathbf{y})$. That is, the set of vectors  $\mathbf{U}$ is used to train either the RBM or the DBN that models $p(\theta)$, the sparse codes $\mathbf{A}$ are used to estimate the variances $\sigma_{{s}_i}^2, \forall i$, using \eqref{eq6.7}, and $\mathbf{E}$ is used to estimate $\sigma_{s_i}^2, \forall i$, using \eqref{luisilla}. For the selection of the generative model configuration, the same procedure described for overcomplete dictionaries in Section \ref{sec:num_hidden} is employed.

Unlike the case of overcomplete dictionaries, the training stage does not require optimization of the sensing matrix. For the case of orthonormal bases, the entries of the sampling matrix $\mathbf{\Phi} \in \mathbb{R}^{M\times N}$ are independently sampled from a normal distribution with mean zero and variance $1/M$.

\section{Experimental Results}
To validate the proposed compressed sensing schemes, a set of experiments are conducted with synthetic and real signals. Results are presented for averages of 50 repetitions of each experiment, with a different realization of the measurement matrix at each iteration. The reconstruction SNR (R--SNR) and the peak SNR (PSNR) are employed as performance measures for one-dimensional signals and images, respectively. The reconstruction SNR is defined as
\begin{equation}\label{BPD1807}
\text{R--SNR (dB)}=10\text{log}_{10}\frac{\|\mathbf{x}\|_2^2}{\|\mathbf{x}-\hat{\mathbf{x}}\|_2^2},
\end{equation}
where $\mathbf{x}$ and $\hat{\mathbf{x}}$ denote the $N$-dimensional original and reconstructed signals, respectively. The PSNR is defined as
\begin{equation}\label{BPD1807}
\text{PSNR (dB)}=10\text{log}_{10}\frac{R^2}{MSE},
\end{equation}
where $R$ is the maximum possible pixel value of the image (255 for 8--bit images) and MSE denotes the mean--square error defined as $\text{MSE}=\frac{1}{N}\|\mathbf{I}-\hat{\mathbf{I}}\|_F^2$, where $\mathbf{I}$ and $\hat{\mathbf{I}}$ denote the original and reconstructed images of size $\sqrt{N}\times \sqrt{N}$, respectively.

In the experiments presented in this section, the proposed methods are compared with the oracle estimate in order to illustrate the best achievable reconstruction. An oracle reveals the true support set $\theta$. Then, such support is used to obtain the estimate of the sparse representation using \eqref{eq6.18}.

In~\cite{Dobi07, Dobi08, Wei15}, a number of samples equal to 500 leads to a good approximation of the empirical average using sampling-based methods.  Following the same choice, the number of samples $H$ for the estimation of the probability distribution of the DBN visible layer is set to 500 in our experiments.

\subsection{Experiments with synthetic signals}
The first set of experiments assume that the parameters of the RBM and DBN are known. The motivation of these experiments is to prevent errors in the parameter estimation, which takes place during training, to propagate into the reconstruction algorithm and, therefore, attain a more faithful evaluation of the algorithm. 

For the first experiment, a set of 50000 sparsity patterns from an RBM, whose parameters are known, are generated via Gibbs sampling. From this set, 45000 and 5000 sparsity patterns are selected at random for training and testing, respectively. The number of hidden units is set equal to the number of visible units. Synthetic signals of dimension $N=256$ are formed with the sparsity patterns and the magnitude of each nonzero coefficient is drawn from a Gaussian distribution with zero mean and known variance $\sigma_{{s}_i}^2\in[10, 50]$. The weights and hidden bias terms are drawn from a uniform distribution, $\{{b}_{{h}_j}, {W}_{ij}\}\sim\mathcal{U}(-1,1), \forall i,j$. The visible bias terms are set to $-14$ to enforce sparsity. A Gaussian sensing matrix $\mathbf{\Phi}$ is employed to sample the testing dataset. Gaussian noise of variance $\sigma_n^2=1$ is added to the measurements. Note that as the synthetic signals are already sparse, a sparsifying transform is not necessary, or equivalently, $\mathbf{D}=\mathbf{I}$. 

The proposed algorithms, RBM--OMP--like and DBN--OMP--like, are executed on the random measurements of the testing dataset and the results are compared with the traditional OMP and with the oracle estimate \eqref{eq6.18}. The RBM-OMP-like algorithm uses the same parameters of the RBM employed to generate the sparsity patterns. The DBN--OMP--like algorithm uses a 2-layer DBN model whose parameters are estimated using the training dataset, as described in Section~\ref{sssec:modeling}. The number of hidden units per layer of the DBN is set the same as the number of visible units. 

\begin{figure}[t]
\centering{ % \hfill
\includegraphics[trim=0 25 0 12, clip, width = \columnwidth]{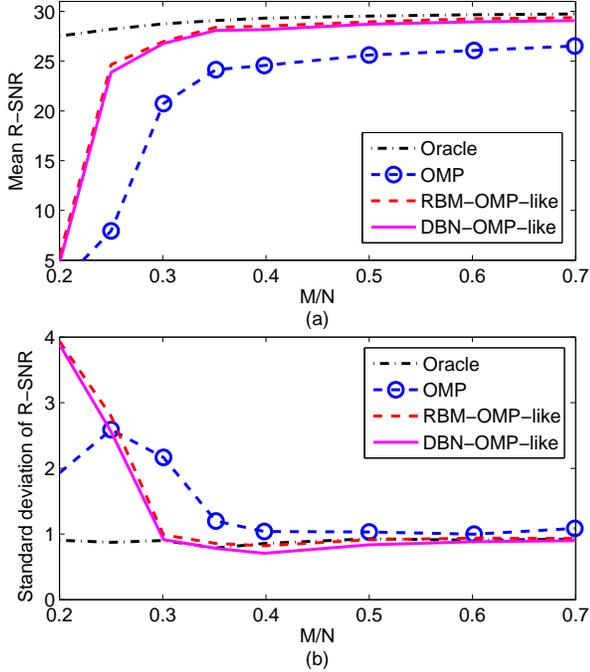}}
\caption{Comparison of the reconstruction of synthetic signals generated with the RBM model using OMP, the RBM-OMP-like algorithm, the DBN-OMP-like algorithm, and the oracle estimator.} \label{res1}
\end{figure}

The mean R-SNR across samples of the testing dataset are shown in Fig.~\ref{res1}(a). Performance gains are observed when using the proposed algorithms. Indeed, their performance is very close to that of the oracle estimate for $M>0.35N$. The proposed algorithms require fewer measurements than the traditional OMP to achieve successful reconstruction. The RBM-–OMP-–like algorithm slightly outperforms the DBN–-OMP-–like algorithm since the data is generated from a probabilistic model that exactly matches the one used by the RBM–-OMP–-like algorithm. Figure~\ref{res1}(b) illustrates the standard deviation of the R-SNR across samples of the testing dataset. The proposed algorithms have lower R-SNR standard deviation than OMP for $M/N>0.25$, which corresponds to their measurement range of successful reconstruction.
%, which is more notorious for low--measurement regimes. 

\begin{figure}[t]
\centering{ % \hfill
\includegraphics[trim=0 25 0 12, clip, width = \columnwidth]{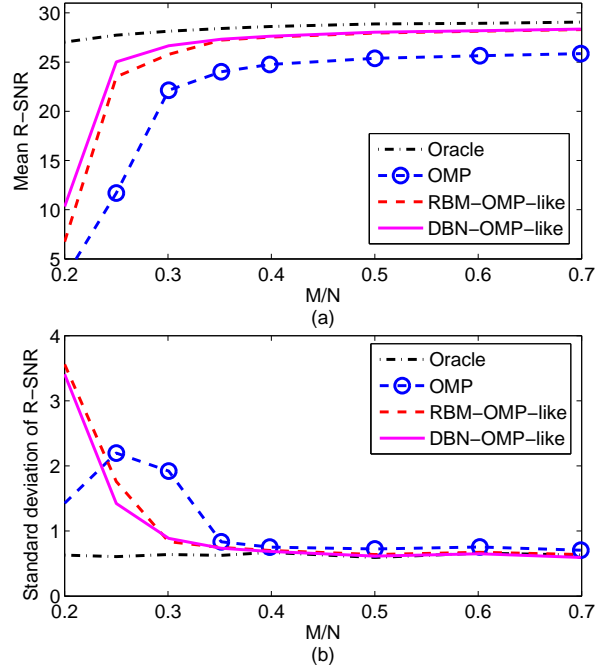}}
\caption{Comparison of the reconstruction of synthetic signals generated with the DBN model using OMP, the RBM-OMP-like algorithm, the DBN-OMP-like algorithm, and the oracle estimator.} \label{res2}
\end{figure}

A similar experiment is conducted with 50000 signals generated from a 2--layer DBN, whose parameters are known \emph{a priori}. The number of hidden units per layer is set the same as the number of visible units. The weights and hidden bias terms are drawn from a uniform distribution, $\{{b}_{{h}^1_j}, {W}^1_{ij}, {b}_{{h}^2_j}, {W}^2_{jk}\}\sim\mathcal{U}(-1,1), \forall i,j,k$. The visible bias terms are set to $-6.5$ to enforce sparsity. As in the previous experiment, 45000 and 5000 sparsity patterns are selected at random for training and testing, respectively. The sampling noise variance and the variance of the nonzero coefficients are kept the same as in the previous experiment, $\sigma_{n}^2=1$ and $\sigma_{{s}_i}^2\in[10, 50], \forall i$, respectively. 

In this experiment, the  proposed algorithms, traditional OMP, and the oracle are employed for reconstruction. The DBN-OMP-like algorithm uses the same parameters of the DBN employed to generate the sparsity patterns. The RBM--OMP--like algorithm uses an RBM model whose parameters are estimated using the training dataset, as described in Section~\ref{sssec:modeling}. The  number  of  hidden  units of the RBM is set the same as the number of visible units. Figure~\ref{res2}(a) indicates that the proposed algorithms outperform OMP, particularly for low--measurement regimes, which demonstrates that exploiting the structural information of the signal's sparsity pattern has the ability to boost the reconstruction performance. 

For example, at a measurement rate of $M = 0.3N$, the DBN--OMP--like algorithm provides a  4.5 dB mean PSNR improvement in performance over OMP. The results are favorably biased towards the DBN--OMP--like algorithm as it uses a distribution that exactly matches the one used to generate the data. The results in Fig.~\ref{res2}(b) indicate that the proposed algorithms only have higher R-SNR standard deviation than OMP for very low--measurement regimes, $M/N<0.23$. 

\begin{table*}[t]
\footnotesize
\renewcommand{\arraystretch}{1.3}
\caption{Reconstruction SNR for different sampling noise variances, $\sigma_{n}^2$.}
\label{table_example}
\centering
\begin{tabular*}{0.85\textwidth}{@{\extracolsep\fill}c|c|cccccccccc}
%\begin{tabularx}{0.8\textwidth}{c|c|cccccccccc}
\hline
\multirow{2}{2cm}{Model used to generate dataset}&\multirow{2}{1.5cm}{Reconstruction algorithm}&\multicolumn{10}{c}{$\sigma_{n}^2$}\\
\cline{3-12}
&&0.5&0.6&0.7&0.8&0.9&1&1.5&2&2.5\\
\hline
\multirow{3}{2.3cm}{RBM-based model}& RBM-OMP-like &31.19&30.32&29.38&29.26&28.94&  28.52  &28.39&26.96 &24.65&\\
&DBN-OMP-like&30.89&30.02&29.08&28.95& 28.71& 28.18 & 28.09&26.78&23.87&\\
&OMP & 28.72& 27.62&26.53&26.08&25.61&   24.57 &24.14&20.74&7.95&\\

\hline
\multirow{3}{2.3cm}
{DBN-based model}&RBM-OMP-like & 29.7 &28.77  & 28.29 & 28.14 & 27.96 &   27.54    & 27.24 & 26.59 & 24.61& \\
&DBN-OMP-like & 29.81 & 28.85&28.36 &28.18 & 28.04&   27.62   &27.33& 26.66 &25.02 &\\
&OMP & 27.7  &  26.75 & 25.86  & 25.64  & 25.39 &   24.76  & 24.02  &22.14   & 11.69  & \\
\hline
\end{tabular*}
\end{table*}

%\begin{table*}[t]
%\footnotesize
%\renewcommand{\arraystretch}{1.3}
%\caption{Reconstruction SNR for different sampling noise variances, $\sigma_\mathbf{n}^2$.}
%\label{table_example}
%\centering
%\begin{tabular*}{0.85\textwidth}{@{\extracolsep\fill}c|c|cccccccccc}
%\begin{tabularx}{0.8\textwidth}{c|c|cccccccccc}
%\hline
%\multirow{2}{2cm}{Model used to generate data set}&\multirow{2}{1.5cm}{Reconstruction algorithm}&\multicolumn{10}{c}{$\sigma_\mathbf{n}^2$}\\
%\cline{3-12}
%&&0.5&0.6&0.7&0.8&0.9&1&1.5&2&2.5&3\\
%\hline
%\multirow{3}{2.3cm}{RBM-based model}& RBM-OMP-like&31.8&30.93&29.99&29.45&28.91&28.09&26.63& 25.44&24.18&23.48\\
%&DBN-OMP-like&&&&&&24.77&&&&\\
%&OMP&28.95&27.85&26.76&26.17&25.6&24.77&23.11&21.2&19.96&18.82\\
%\hline
%\multirow{3}{2.3cm}{DBN-based model}&DBN-OMP-like & 28.98 &28.02 &27.53 &26.79 &26.3&25.71&23.89& 22.44 &21.23 &20.32\\
%&OMP &  26.63 &  25.68 &  24.79 &  23.94 & 23.49 &  22.81 &  20.51 &  18.76 &  17.64 & 16.42\\
%\hline
%\end{tabular*}
%\end{table*}

Next, we demonstrate that the proposed algorithms are stable in the presence of measurement noise using the same datasets as those of the previous experiments. For this experiment, the number of measurements is set to $M=0.4N$, as Figs.~\ref{res1} and~\ref{res2} indicate that the reconstruction is successful when such a number of measurements is used. The Gaussian sensing matrix is kept fixed while  the sampling noise variance is varied in the range $[0.5, 2.5]$. Table~\ref{table_example} indicates that both RBM--OMP--like and DBN--OMP--like algorithms outperform the traditional OMP algorithm for the specified range of variances. Below $\sigma_{n}^2=2$, the proposed algorithms produce faithful reconstructions with an SNR greater than 26 dB.

 \subsection{Experiments with the MNIST Database}
In this section, the performance of the proposed reconstruction algorithms as a function of the number of hidden units and hidden layers is studied via numerical experiments. The MNIST dataset~\cite{Lecu98}, which contains 70000 grayscale images of handwritten digits of size $N=28\times28$, is employed for the experiments. The dataset is divided into 40000 samples for training, 20000 samples for validation and 10000 samples for testing. Since the images are already sparse in the spatial domain, a sparsifying transform is not necessary, or equivalently, $\mathbf{D}=\mathbf{I}$. The validation and testing datasets are sampled with a matrix $\mathbf{\Phi}$ whose entries are drawn from a zero--mean Gaussian distribution with variance $1/M$. The resulting compressed measurements are artificially contaminated with Gaussian noise of variance $\sigma_{n}^2=1.2$. 

\begin{figure}[t]
\centering{ % \hfill
\includegraphics[width = \columnwidth]{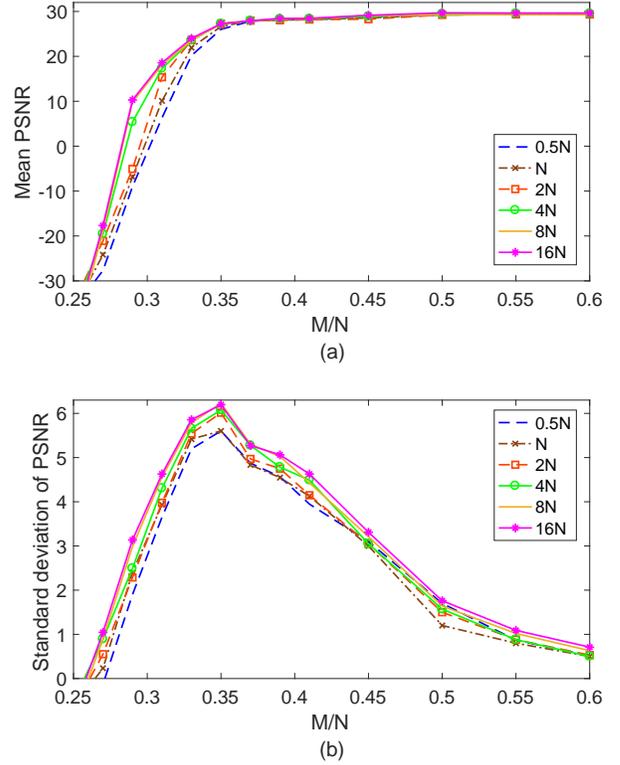}}
\caption{Evaluation of the MNIST validation dataset reconstruction using RBM models with different number of hidden units.} \label{figina}
\end{figure}

As described in Section~\ref{sec:num_hidden}, the optimal model configurations are selected via cross-validation. RBM and DBN models with different number of hyperparameters are learned using the training dataset and evaluated using the validation dataset. Figure~\ref{figina}(a) illustrates the mean PSNR across samples of the validation dataset as a function of the number of measurements for different RBM models, which are employed by the RBM--OMP--like algorithm. The reconstruction performance improves as the number of hidden units of the RBM model increases from $0.5N$ to $8N$. The standard deviation across samples of the validation dataset also increases as the number of hidden units increases (Fig.~\ref{figina}(b)). The reconstruction curves in Fig.~\ref{figina}(a) corresponding to RBMs with $8N$ and $16N$ hidden units are very similar, which may indicate that their representational power is almost the same. Therefore, setting the number of hidden units above $8N$ does not improve the reconstruction performance and may lead to overfitting. Consequently, we select an RBM with $8N$ hidden units for testing.

Tramel \textit{et al}. also evaluated their RBM-based reconstruction method using the MNIST database~\cite{Trame15}. They employed the percent of successfully recovered digit images as performance metric instead of the mean R-SNR. They showed that they could successfully recover 90\% digit images from their testing dataset with only $0.25N$ measurements, while we attain poor reconstruction performance when using only $0.25N$ measurements as it is shown in Fig.~\ref{figina}(a). However, this is not a fair comparison since the authors of ~\cite{Trame15} selected different values for the model and experiment parameters. For example, they selected a noise variance that was much smaller than ours, a larger training dataset and a smaller testing dataset. The RBM--OMP--like algorithm is expected to improve the reconstruction performance of the MNIST dataset when noise of smaller variance is added to the signals.

\begin{figure}[t]
\centering{ % \hfill
\includegraphics[width = \columnwidth]{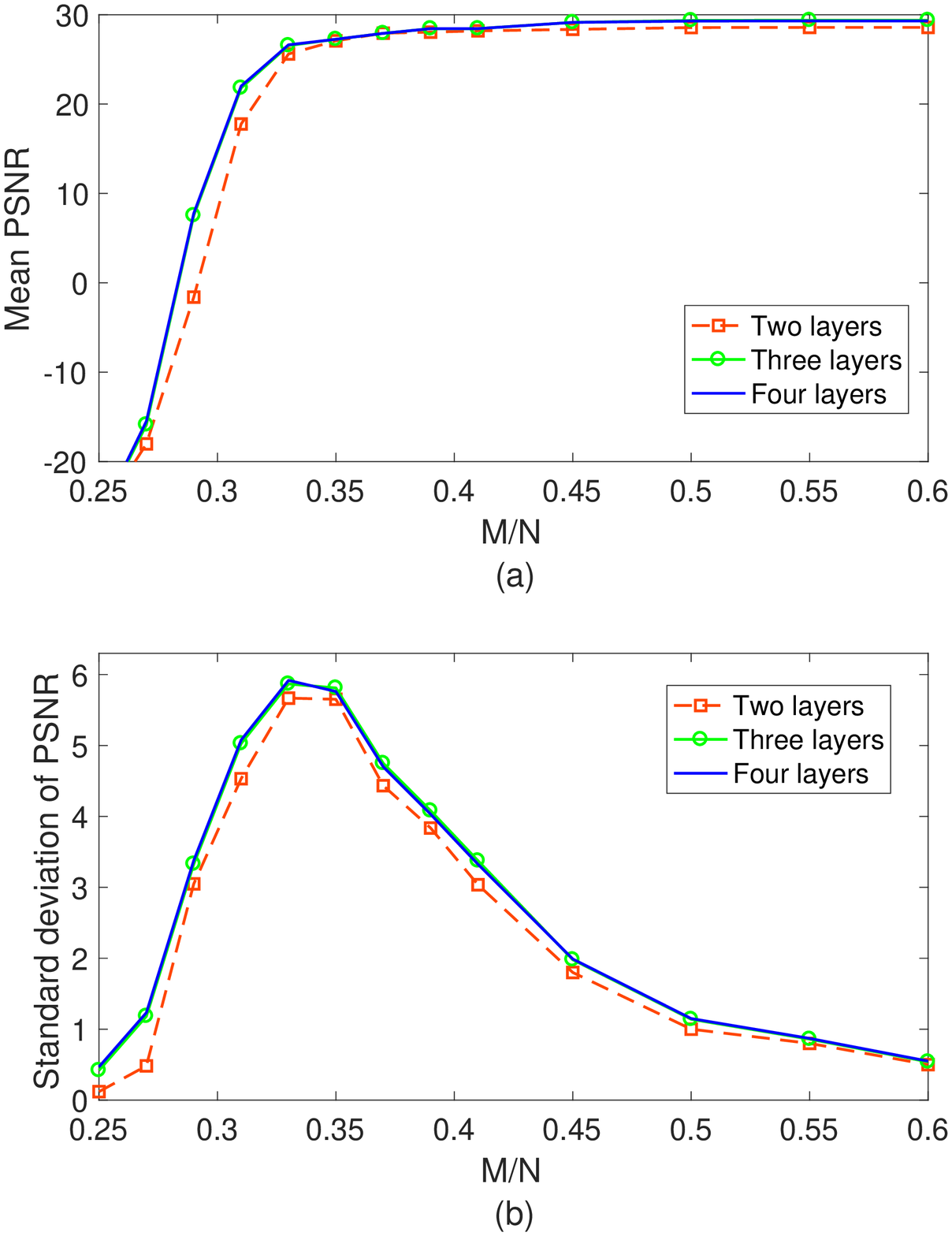}}
\caption{Evaluation of the MNIST validation dataset reconstruction using DBN models with different number of hidden layers.} \label{figina2}
\end{figure}

A similar experiment is performed to select the number of hidden layers of the DBN employed by the DBN--OMP--like algorithm. Deep and narrow belief networks, with hidden layers of the same dimension as that of the visible layer, are considered. Figures~\ref{figina2}(a) and~\ref{figina2}(b) illustrate that the mean PSNR and standard deviation across samples of the validation dataset increase as the number of hidden layers changes from $L=2$ to $L=3$, respectively. Setting the number of hidden layers above 3 does not improve the reconstruction performance, which suggests that the representational power of the DBN does not improve for $L>3$. Therefore, we select a DBN with 3 hidden layers, whose number of hidden units per layer is the same as the number of visible units, for testing. By comparing Figs.~\ref{figina} and \ref{figina2}, it is noted that using an RBM with 8 hidden units leads to a similar performance as using a DBN with three hidden layers. However, the DBN architecture requires fewer parameters to be trained. More precisely, the RBM model requires $8N^2+9N=4924304$ parameters ($8N^2$ weights and $9N$ bias terms) while the DBN model only requires $3N^2+4N=1847104$ parameters ($3N^2$ weights and $4N$ bias terms). This comparison illustrates that RBMs and DBNs have the same representational power, but DBNs may lead to more compact representations.  

\begin{figure}[t]
\centering{ % \hfill
\includegraphics[width = \columnwidth]{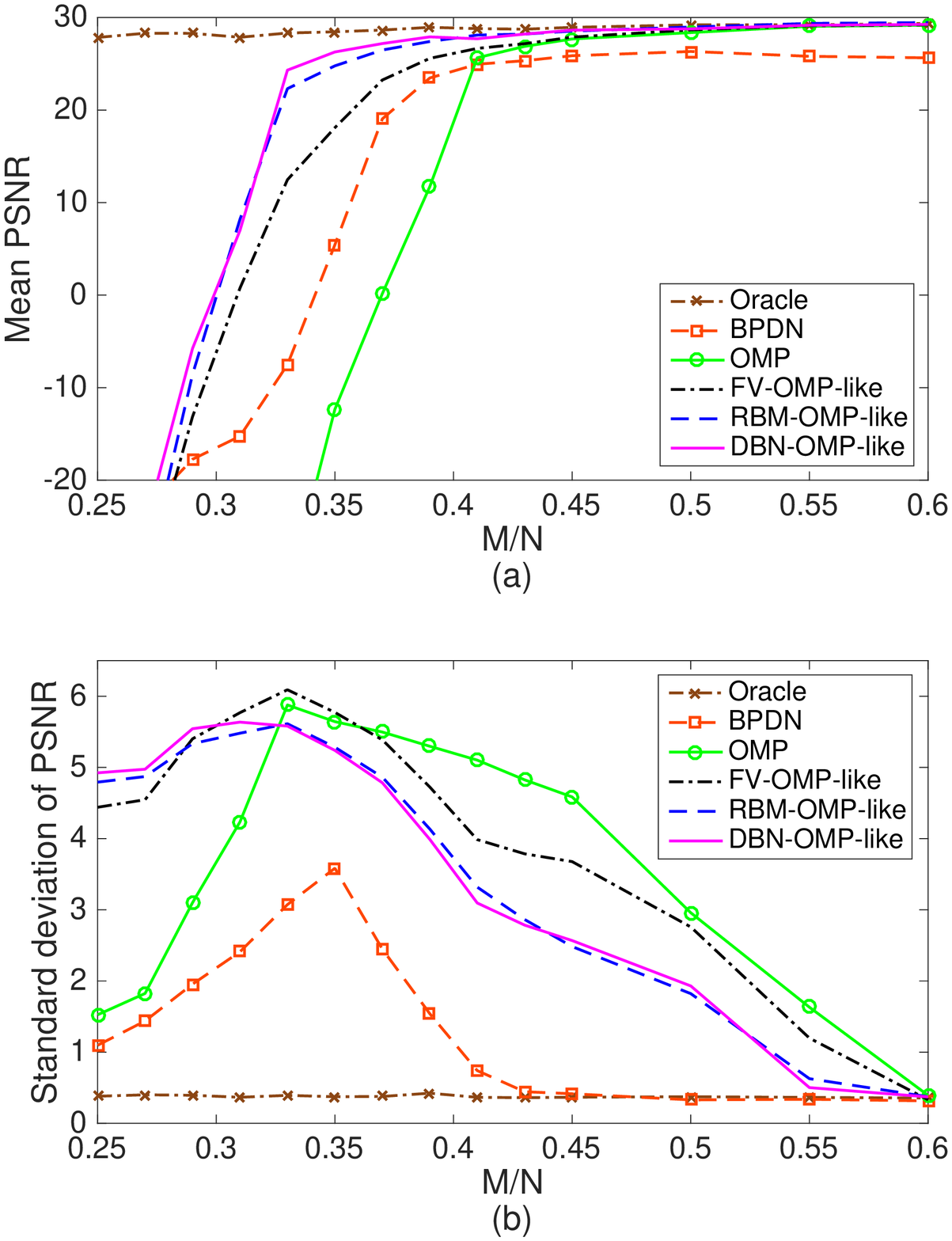}}
\caption{Evaluation of the reconstruction of images from the MNIST dataset. The OMP, BPDN, FV-OMP-like, RBM-OMP-like, DBN-OMP-like, and the oracle are employed as reconstruction algorithms.} \label{figina3}
\end{figure}

Figures~\ref{figina3} compares the performance of the proposed algorithms on the testing dataset, using the model configurations chosen via cross-validation, with CS reconstruction algorithms, such as OMP and basis pursuit denoising (BPDN)~\cite{Chen98}. We also compare with the MAP--OMP--like algorithm presented in~\cite{Pele12} (henceforth referred to as FV--OMP-like algorithm), which has the same structure as the RBM--OMP--like algorithm, with the difference being that it employs a fully visible Boltzmann machine to model the probability distribution of the sparsity pattern. The proposed algorithms attain the best reconstruction performance, followed closely by the FV--OMP--like algorithm, which also achieves successful reconstruction given that it also exploits statistical dependencies in the sparsity pattern. Algorithms that do not exploit structural information beyond sparsity require more measurements to achieve successful reconstruction than the proposed algorithms. For example, OMP and BPDN require $M>0.41N$ to attain $\text{PSNR}>25$ dB. Contrarily, the proposed algorithms only require $M>0.35N$ measurements to attain $\text{PSNR}>25$ dB. On the downside, the proposed algorithms have higher PSNR standard deviation than BPDN as shown by Fig.~\ref{figina3}(b). 

\subsection{Experiments with the Berkeley  segmentation  dataset}
This section presents experimental validation of the proposed algorithms using real data from the Berkeley segmentation dataset~\cite{Mart01}. This dataset contains 400 images of real--life scenes.  Each image is comprised of $321\times481$ pixels. Experiments are conducted with the proposed compressed sensing schemes using orthonormal bases and overcomplete learned dictionaries as sparsifying transforms. The training dataset for these experiments consists of 50400 patches of size $N=8 \times 8$, extracted at random from a set of 350 images. Similarly, the testing dataset consist of 2000 $8 \times 8$ patches, extracted at random from a different set of 50 images. The same number of patches is extracted from each image. 

An RBM with the same number of hidden units as visible units and a 2--layer DBN are employed to model the probability distribution of the sparsity pattern. The number of hidden units per layer of the DBN is set to half the number of visible units. Compressed measurements are artificially contaminated with Gaussian noise of variance $\sigma_{n}^2=1$. In the case of orthonormal bases, the entries of the sensing matrix $\mathbf{\Phi}$ are drawn from a zero--mean Gaussian distribution with variance $1/M$. In the case of overcomplete learned dictionaries, the sensing matrix is optimized as described in~\cite{Duar09} and the number of dictionary atoms is set to $2N$. The sparsity level for overcomplete dictionaries is set to $K=0.15N$.

The scheme illustrated in Fig.~\ref{block1} is considered first. Since the wavelet transform has proven useful for compressing natural images, the Symlets--8 wavelet transform, using a decomposition level $L=4$, is chosen as the orthonormal sparsifying transform for our experiments. The proposed algorithms, RBM--OMP--like and DBN--OMP--like, are compared with the oracle estimator, OMP, and BPDN. We also compare with other algorithms that exploit structural information beyond sparsity, such as the Bayesian compressed sensing algorithm described in~\cite{He10} (henceforth referred to as TS-BCS)  and the FV--OMP-like algorithm, which has the same structure as the RBM--OMP--like algorithm, with the difference being that it employs a fully visible Boltzmann machine to model the probability distribution of the sparsity pattern. 

\begin{figure}[t]
\centering{ % \hfill
\includegraphics[width = \columnwidth]{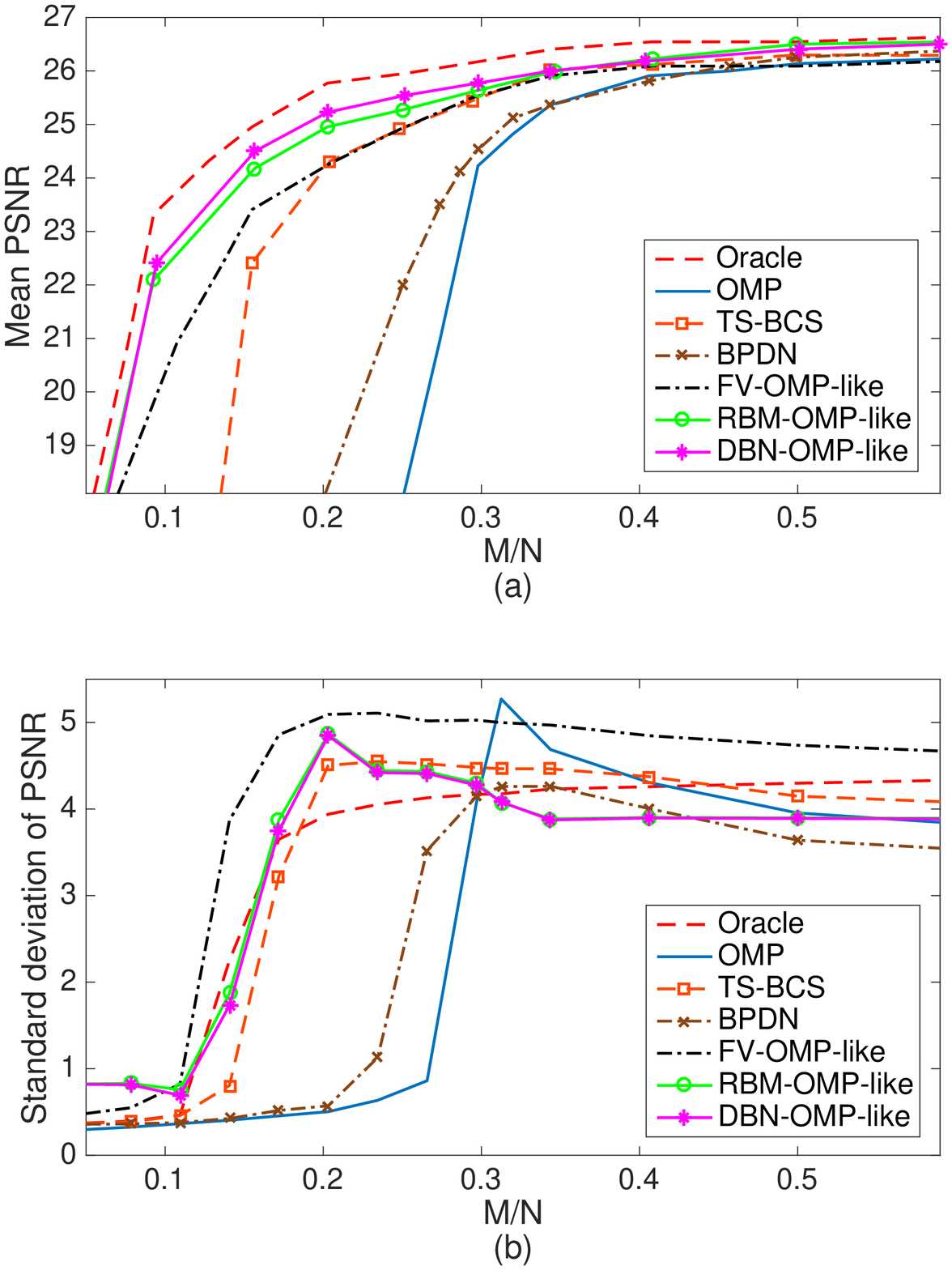}}
\caption{Evaluation of the reconstruction of natural images from random projections using wavelets as the sparsifying transform. The oracle estimator, OMP, BPDN, TS-BCS, FV-OMP-like, RBM-OMP-like, and DBN-OMP-like are employed as reconstruction algorithms.} \label{res5}
\end{figure}

\begin{figure}[t]
\centering{ % \hfill
\includegraphics[width = \columnwidth]{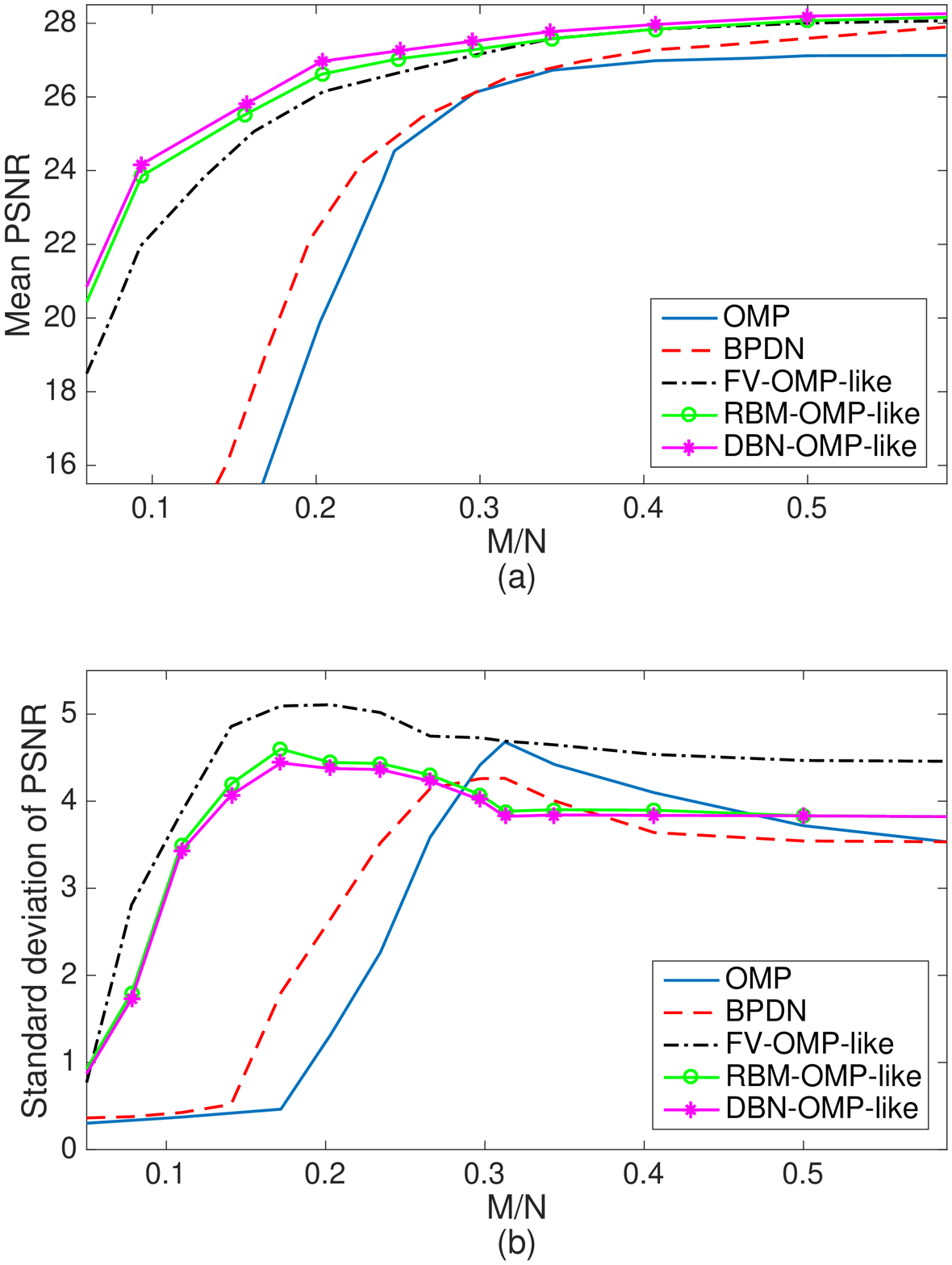}}
\caption{Evaluation of the reconstruction of natural images from random projections using overcomplete learned dictionaries as the sparsifying transform. Dictionary training algorithms DL1 and DL2 are employed. The OMP, BPDN, FV-OMP-like, RBM-OMP-like, and DBN-OMP-like are employed as reconstruction algorithms.} \label{res6}
\end{figure}

In this experiment, the true support revealed by the oracle corresponds to the best $K$--term approximation of the signal to be recovered. As in our proposed algorithms, the FV--OMP--like algorithm requires the parameters of the support prior distribution. Such parameters are learned from the training dataset using the maximum likelihood approach described in~\cite{Pele12}.

\begin{figure*}[t]
\centering{ % \hfill
\hspace*{-25pt} \includegraphics[width =.95\textwidth]{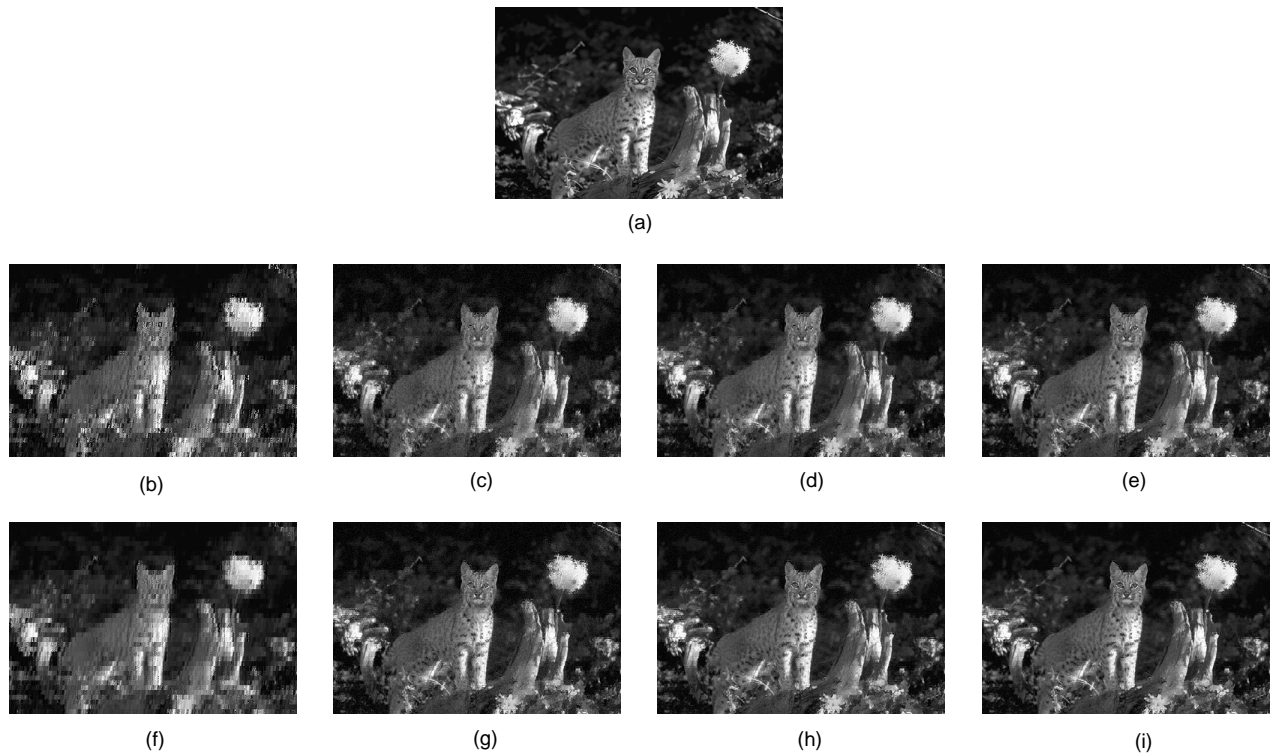}}
\caption{Visual evaluation of the proposed reconstruction algorithms using overcomplete dictionaries and the wavelet basis as sparsifying transforms. $M=0.2N$. First row: (a) Original image. Second row: (b-e) Reconstructed images using a wavelet basis as sparsifying transform (b) OMP reconstruction, PSNR=17.09, (c) FV-OMP-like reconstruction, PSNR=22.7, (d) RBM-OMP-like reconstruction, PSNR=24.19, (e) DBN-OMP-like reconstruction, PSNR=25.04. Third row: (f-i) Reconstructed images using an overcomplete learned dictionary as sparsifying transform, (f) OMP reconstruction, PSNR=20.11, (g) FV-OMP-like reconstruction, PSNR=24.28, (h) RBM-OMP-like reconstruction, PSNR=25.82, (i) DBN-OMP-like reconstruction, PSNR=27.1.} \label{fig:visual1}
\end{figure*}

The results of the comparison in terms of mean PSNR across the testing dataset are reported in Fig.~\ref{res5}(a). It is noted that both RBM--OMP--like and DBN--OMP--like algorithms outperform OMP, BPDN, TS-BCS, and the FV--OMP--like algorithm. The number of parameters of the RBM and DBN models are $N^2+2N$ ($N^2$ weights and $2N$ bias terms) and $3N^2/4+2N$ ($3N^2/4$ weights and $2N$ bias terms), respectively. Even though the DBN requires fewer parameters to be learned, the performance of the DBN--OMP--like algorithm is slightly superior to that of the RBM--OMP--like algorithm due to the multi--layer structure of the DBN. As the reconstruction performance is related to the representational power of the model, the results in Fig.~\ref{res5}(a) are consistent with the work by Le Roux \textit{et al}.~\cite{Le08}, which shows that, even though a DBN and an RBM can have the same representational power, a DBN offers a more compact and efficient representation in terms of number of parameters. 

Figure~\ref{res5}(b) illustrates the standard deviation of PSNR across samples of the testing dataset for the different reconstruction algorithms. The FV--OMP--like algorithm exhibits the largest standard deviation for almost the entire measurement range. Compared to the other algorithms, the proposed algorithms exhibit low standard deviation for $M/N>0.3$. 

Consider next the implementation of the scheme illustrated in Fig.~\ref{block2}. In this case, the overcomplete dictionary is learned using the K--SVD algorithm. An experiment is conducted to evaluate the performance of the proposed algorithms and the results of the mean PSNR across the testing dataset are shown in Fig.~\ref{res6}(a). The RBM--OMP--like and DBN--OMP--like algorithms do not only require fewer measurements than conventional OMP to achieve stable recovery but also attain higher mean PSNR values for the entire range of measurements. There is a small performance gap in favor of the DBN--OMP--like algorithm when compared to the RBM--OMP--like algorithm. The proposed algorithms also outperform the FV--OMP--like algorithm, which suggests that exploiting higher--order dependencies between the sparse representation coefficients leads to superior reconstruction performance. As in the previous experiment, the proposed algorithms exhibit lower standard deviation than the FV--OMP--like algorithm (Fig.~\ref{res6}(b)).

Finally, visual evaluation of a reconstructed test image using the proposed CS schemes is presented in Fig.~\ref{fig:visual1}. Each non--overlapping $8\times8$ patch from the image was reconstructed from their noisy projections using $M=2\times N$ measurements and sampling noise of variance $\sigma_{n}^2=1$. The first row of Fig.~\ref{fig:visual1} shows the original image (Fig.~\ref{fig:visual1}(a)). The second and third rows show the results when using wavelets and overcomplete dictionaries as sparsifying transforms, respectively. In the second and third rows, the first, second, third and fourth columns correspond to the reconstruction via the OMP, FV--OMP--like, RBM--OMP--like and DBN--OMP--like algorithms, respectively. For both overcomplete dictionaries and orthonormal bases, the RBM--OMP--like and DBN--OMP--like algorithms generate  higher quality images than the OMP and FV--OMP--like algorithms, as can be noticed by the reduction of artifacts, the sharper edges, and the preservation of details. In contrast, images reconstructed with the OMP algorithm have poor quality as OMP does not exploit any structure beyond sparsity.

%%%%%%%%%%
%%%%%%%%%

\section{Conclusions}
Deep learning is one of the most powerful representation learning techniques. In this paper, it was shown how the ability of one deep learning architecture, the deep belief network, and restricted Boltzmann machines to capture the complex statistical structure of the input data can be leveraged by CS systems. Statistical dependencies are informative and exploiting them leads to improvements in reconstruction performance.

The proposed scheme operates over signals belonging to a certain signal class. In this paper, the signal classes of natural images from the Berkeley Segmentation Dataset and of handwritten digits from the MNIST Database were selected, but the scheme can also be applied to other signal classes; \textit{e.g.} radar, ECG, EEG, medical imaging, speech signals, \textit{etc}. Restricted Boltzmann machines and DBNs were employed to model a prior distribution for the sparsity pattern of the signal class. Such a prior was employed by a MAP estimator for the reconstruction. It was shown through simulations that the proposed scheme leads to significantly superior reconstruction results when compared with CS methods that do not exploit any statistical dependencies between dictionary atoms. The proposed approach also outperforms CS methods that only exploit pair--wise correlations between dictionary atoms.

\bibliographystyle{IEEEbib}
\bibliography{abrev,RGCD2}

\begin{IEEEbiography}[{\includegraphics[width=1in,height=1.25in,clip,keepaspectratio]{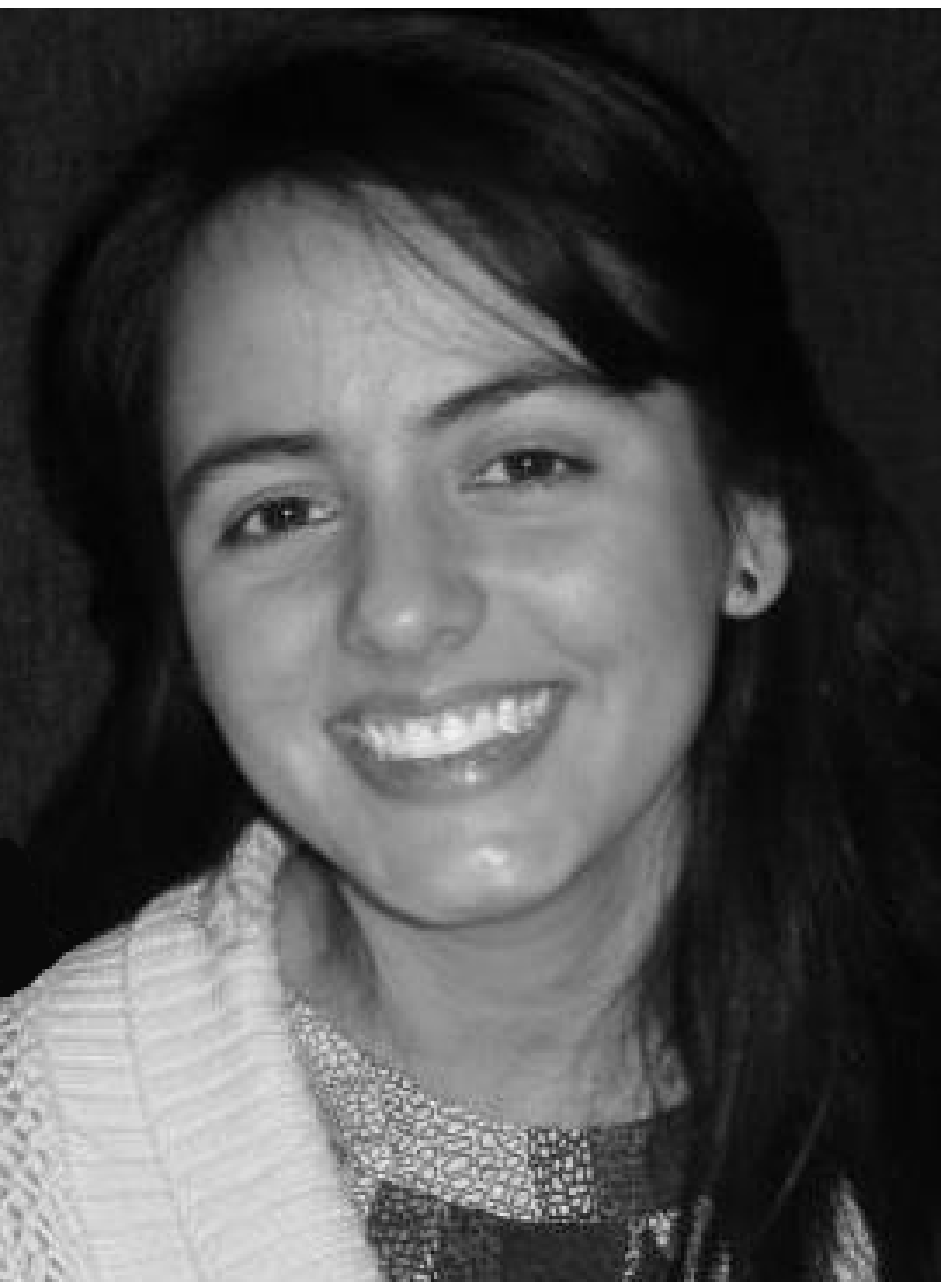}}]{Luisa F. Polan\'{i}a}
(S'12-M'15) received the B.S.E.E. degree (with honors) from the National University of Colombia, Bogot\'{a}, Colombia, in 2009. She received the Ph.D. degree in Electrical and Computer Engineering from the University of Delaware, Newark, Delaware, in 2015. 

After the Ph.D. degree, she held a research position at the Palo Alto Research Center from October 2014 to
June 2016. She is currently a machine learning scientist at American Family Mutual Insurance Company in Madison, WI. Her research interests include signal and image processing, compressive sensing, low-dimensional modeling, and machine learning. Miss Polan\'{i}a was the recipient of the University of Delaware Graduate Student Fellowship in 2013 and the PARC Exceptional Performance Award in 2015.
\end{IEEEbiography}

\begin{IEEEbiography}[{\includegraphics[width=1in,height=1.25in,clip,keepaspectratio]{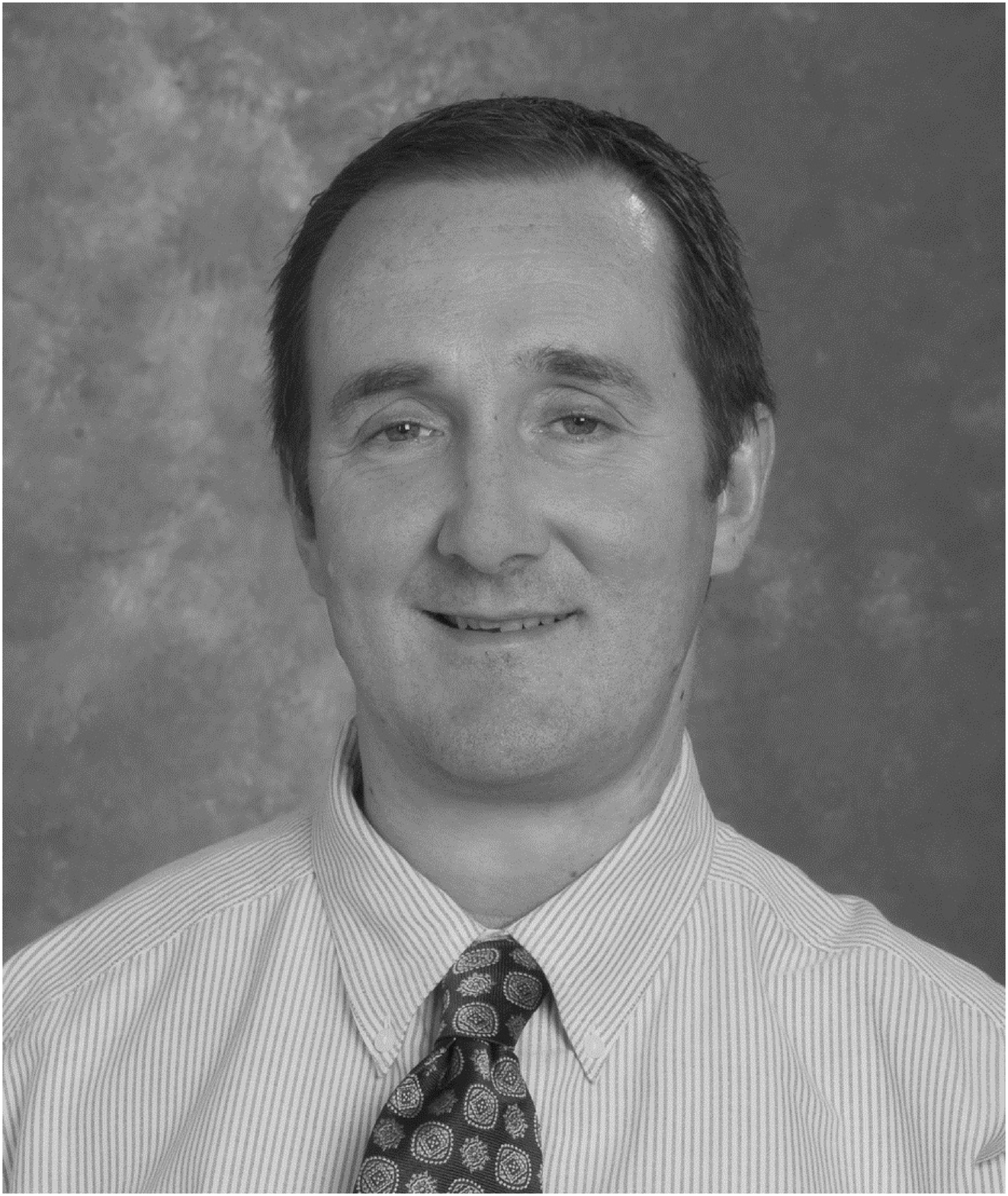}}]{Kenneth E. Barner}
(S'84-M'92-SM'00-F'16) received the B.S.E.E. degree (magna cum laude) from Lehigh University, Bethlehem, Pennsylvania, in 1987. He received the M.S.E.E. and Ph.D. degrees from the University of Delaware, Newark, Delaware, in 1989 and 1992, respectively.

He was the duPont Teaching Fellow and a Visiting Lecturer with the University of Delaware in 1991 and 1992, respectively. From 1993 to 1997, he was an Assistant Research Professor with the Department of Electrical and Computer Engineering, University of Delaware, and a Research Engineer with the duPont Hospital for Children. He is currently Professor and Chairman with the Department of Electrical and Computer Engineering, University of Delaware. He is coeditor of the book \emph{Nonlinear Signal and Image Processing: Theory, Methods, and Applications} (Boca Raton, FL: CRC), 2004. His research interests include signal and image processing, robust signal processing, nonlinear systems, sensor networks and consensus systems, compressive sensing, human-computer interaction, haptic and tactile methods, and universal access.

Dr. Barner is the recipient of a 1999 NSF CAREER award. He was the Co-Chair of the 2001 \emph{IEEE-EURASIP Nonlinear Signal and Image Processing (NSIP) Workshop} and a Guest Editor for a Special Issue of the \emph{EURASIP Journal of Applied Signal Processing} on Nonlinear Signal and Image Processing. He is a member of the Nonlinear Signal and Image Processing Board. He was the Technical Program Co-Chair for ICASSP 2005 and and previously served on the IEEE Signal Processing Theory and Methods (SPTM) Technical Committee and the IEEE Bio-Imaging and Signal Processing (BISP) Technical Committee. He is currently a member of the IEEE Delaware Bay Section Executive Committee. He has served as an Associate Editor of the \emph{IEEE Transactions on Signal Processing}, the \emph{IEEE Transaction on Neural Systems and Rehabilitation Engineering}, the \emph{IEEE Signal Processing Magazine}, the \emph{IEEE Signal Processing Letters}, and the \emph{EURASIP Journal of Applied Signal Processing}. He was the Founding Editor-in-Chief of the journal \emph{Advances in Human-Computer Interaction}. For his dissertation ``Permutation Filters: A Group Theoretic Class of Non-Linear Filters,'' Dr. Barner received the \emph{Allan P. Colburn Prize in Mathematical Sciences and Engineering} for the most outstanding doctoral dissertation in the engineering and mathematical disciplines. He is a member of Tau Beta Pi, Eta Kappa Nu, and Phi Sigma Kappa.
\end{IEEEbiography}

\end{document}